\documentclass[prodmode,acmcsur]{acmstyle}

\usepackage[utf8]{inputenc}
\usepackage{amsmath}
\usepackage{amssymb}
\usepackage{amscd}
\usepackage{amsfonts}
\usepackage{graphicx}
\usepackage{epsfig}
\usepackage{epstopdf}
\usepackage{algorithm}
\usepackage{algorithmic}
\usepackage{float}
\usepackage{url}
\usepackage{color}
\usepackage{multirow}
\usepackage{rotating}
\usepackage{natbib}
\usepackage{booktabs}
\newcommand{\R}[1]{\mathbb{R}^{#1}}
\newcommand{\RR}[2]{\mathbb{R}^{#1 \times #2}}

\newcommand{\rank}[1]{\mbox{rank}(#1)}

\newcommand{\Proj}{\mathcal{P}}

\newcommand{\SVT}{\mathcal{D}}

\newcommand{\st}{\mbox{s.t.~}}

\newcommand{\refEq}[1]{(\ref{#1})}
\newcommand{\refFig}[1]{Figure~\ref{#1}}

\newcommand{\refTheor}[1]{Theorem~\ref{#1}}
\newcommand{\refSec}[1]{Section~\ref{#1}}

\newcommand{\refAlg}[1]{Algorithm~\ref{#1}}
\newcommand{\refTab}[1]{Table~\ref{#1}}

\def\Pr{\text{Pr}}
\def\eg{{e.g.~}}
\def\ie{{i.e.~}}

\def\bfa{{\mathbf{a}}}
\def\bfb{{\mathbf{b}}}

\def\bfd{{\mathbf{d}}}
\def\bfe{{\mathbf{e}}}

\def\bfk{{\mathbf{k}}}

\def\bfu{{\mathbf{u}}}
\def\bfv{{\mathbf{v}}}
\def\bfw{{\mathbf{w}}}
\def\bfx{{\mathbf{x}}}
\def\bfy{{\mathbf{y}}}

\def\bfA{{\mathbf{A}}}
\def\bfB{{\mathbf{B}}}
\def\bfC{{\mathbf{C}}}
\def\bfD{{\mathbf{D}}}
\def\bfE{{\mathbf{E}}}

\def\bfI{{\mathbf{I}}}

\def\bfM{{\mathbf{M}}}

\def\bfP{{\mathbf{P}}}
\def\bfQ{{\mathbf{Q}}}

\def\bfS{{\mathbf{S}}}

\def\bfU{{\mathbf{U}}}
\def\bfV{{\mathbf{V}}}

\def\bfX{{\mathbf{X}}}
\def\bfY{{\mathbf{Y}}}
\def\bfZ{{\mathbf{Z}}}

\def \bfalpha {\boldsymbol{\alpha}}

\def \bfgamma {\boldsymbol{\gamma}}

\def\bfPhi{{\mathbf{\Phi}}}

\def\half{\frac{1}{2}~}
\def\reg{\mathcal{R}}

\newcommand{\hl}[1]{{#1}}

\acmVolume{0}
\acmNumber{0}
\acmArticle{0}
\acmYear{2014}
\acmMonth{9}

\begin{document}

% Page heads
\markboth{X. Zhou et al.}{Low-Rank Modeling and Its Applications in Image Analysis}

\title{Low-Rank Modeling and Its Applications in Image Analysis}
\author{Xiaowei Zhou
\affil{The Hong Kong University of Science and Technology}
Can Yang
\affil{Hong Kong Baptist University}
Hongyu Zhao
\affil{Yale University}
Weichuan Yu
\affil{The Hong Kong University of Science and Technology}
}
%\date{}

\begin{abstract}
Low-rank modeling generally refers to a class of methods that solve problems by representing variables of interest as low-rank matrices. It has achieved great success in various fields including computer vision, data mining, signal processing and bioinformatics. Recently, much progress has been made in theories, algorithms and applications of low-rank modeling, such as exact low-rank matrix recovery via convex programming and matrix completion applied to collaborative filtering. These advances have brought more and more attentions to this topic. In this paper, we review the recent advance of low-rank modeling, the state-of-the-art algorithms, and related applications in image analysis. We first give an overview to the concept of low-rank modeling and challenging problems in this area. Then, we summarize the models and algorithms for low-rank matrix recovery and illustrate their advantages and limitations with numerical experiments. Next, we introduce a few applications of low-rank modeling in the context of image analysis. Finally, we conclude this paper with some discussions.
\end{abstract}

%\category{I.5.2}{Pattern Recognition}{Design Methodology}

%\terms{Algorithms}

%\keywords{Low-rank modeling, matrix factorization, optimization, image analysis}

%\acmformat{Xiaowei Zhou, Can Yang, Hongyu Zhao and Weichuan Yu, 2014. Low-Rank Modeling and Its Applications in Image Analysis.}

% At a minimum you need to supply the author names, year and a title.
% IMPORTANT:
% Full first names whenever they are known, surname last, followed by a period.
% In the case of two authors, 'and' is placed between them.
% In the case of three or more authors, the serial comma is used, that is, all author names
% except the last one but including the penultimate author's name are followed by a comma,
% and then 'and' is placed before the final author's name.
% If only first and middle initials are known, then each initial
% is followed by a period and they are separated by a space.
% The remaining information (journal title, volume, article number, date, etc.) is 'auto-generated'.

%\begin{bottomstuff}
%This work was partially supported by a grant from the Research Grant Council of the Hong Kong Special Administrative Region, China (Project No: T12-402/13-N), by NIH grants GM59507 and CA154295, and by NSF grant DMS1106738.
%
%Author's addresses: X. Zhou {and} W. Yu, Department of Electronic and Computer Engineering,
%The Hong Kong University of Science and Technology, Hong Kong, China;
%C. Yang, Department of Mathematics, Hong Kong Baptist University, Hong Kong, China;
%H. Zhao, Department of Biostatistics, School of Public Health, Yale University, New Haven, CT 06520.
%\end{bottomstuff}

\maketitle

\section{Introduction}

In many research fields the data to be analyzed often have high dimensionality, which brings great challenges to data analysis. Some examples include images in computer vision, documents in natural language processing, customers' records in recommender systems, and genomics data in bioinformatics.

Fortunately, the high-dimensional data often lie in a low-dimensional subspace. Mathematically, if we represent each data point by a vector $\bfd_i\in\mathbb{R}^m$ and denote the entire dataset by a big matrix $\bfD=\left[\bfd_1,\cdots,\bfd_n\right]$, the low-dimensionality assumption can be translated into the following low-rank assumption: $\rank{\bfD}\ll\min(m,n)$. A typical example in computer vision is Lambertian reflectance, where $\bfd_1,\cdots,\bfd_n$ correspond to a set of images of a convex Lambertian surface under various lighting conditions \citep{basri2003lambertian}. Another example is from signal processing, where $\bfd_i$ represents a vector of signal intensities received by an antenna array at time point $i$. Interested readers are referred to \citep[Chapter 1.3]{markovsky12lowrank} for more low-rank examples.

In the real world, the raw data can hardly be perfectly low-rank due to the existence of noise. Therefore, the following model is more faithful to real situations
\begin{align}
\bfD = \bfX + \bfE,
\end{align}
where $\bfX$ corresponds to a low-rank component and $\bfE$ corresponds to the noise or error in measurements. Recovering the low-rank structure from noisy data becomes the centric task in many problems.

A conventional approach to finding the low-rank approximation is to solve the following optimization problem:
\begin{align}\label{eq:mafit_l2}
\min_{\bfX}~ &\|\bfD-\bfX\|_F^2, \nonumber \\
\st &\rank{\bfX}\leq r,
\end{align}
where $\|\bfY\|_F=\sqrt{\sum_{ij}{Y_{ij}^2}}$ denotes the Frobenious norm of a matrix.\footnote{In this paper, a matrix is denoted by a capital letter, e.g. $\bfY$. An element and a column of $\bfY$ are denoted by $Y_{ij}$ and $\bfy_i$, respectively.} Solving such a minimization problem can be interpreted as seeking the optimal rank-$r$ estimate of $\bfD$ in a least-squares sense. According to the matrix approximation theorem \citep{eckart1936approximation}, the solution to \refEq{eq:mafit_l2} is given analytically by the singular value decomposition (SVD):
\begin{align}\label{eq:mal}
\bfX^* = \sum_{i=1}^{r}{\sigma_i\bfu_i\bfv_i^T},
\end{align}
where $\{\bfu_i\}$, $\{\bfv_i\}$, and $\{\sigma_i\}$ for $i=1,\cdots,r$ correspond to the left singular vectors, right singular vectors, and singular values of $\bfD$, respectively. The vectors $\bfu_1,\cdots,\bfu_r$ also form an orthogonal basis to represent a $r$-dimensional subspace that can best embed the data. This procedure corresponds to Principal Component Analysis (PCA) \citep{jolliffe2002principal} in statistics.

While PCA is one of the most popular tools for data analysis because of the analytical solution in computation and the provable optimality under certain assumptions, it cannot handle some difficulties in real applications. Consider the following two common examples:

\vspace{1em}
\noindent{\bf Recovery from a few entries.} In many applications, we would like to recover a matrix from only a small number of observed entries. A typical example is that, when building recommender systems, we hope to make predictions to customers' preferences based on the information collected so far. The NetFlix problem \citep{koren2009matrix} is a famous instance. The data is a big matrix $\bfD$ with each entry $D_{ij}\in\{1,\cdots,5\}$ recording the rating of customer $i$ for movie $j$. There are around 480K customers and 18K movies in the dataset, but only $1.2\%$ entries have values since each customer only rated about 200 movies on average. The problem is how to predict the ratings that have not been made yet based on the current observation. A popular solution is to assume that the rating matrix should be low-rank. This assumption is based on the fact that a subgroup of customers are likely to share a similar taste and their ratings to the movies will be highly correlated. Consequently, the rank of the rating matrix will be bounded by the number of subgroups formed by the customers. Therefore, the problem turns into recovering a low-rank matrix from a few entries. This problem is often called {\bf Matrix Completion} (MC).

\vspace{1em}
\noindent{\bf Recovery from gross errors.} In some other applications, we have to recover a low-dimensional subspace from corrupted data. For example, the face images of a person may include glasses or shadows that occlude the true appearance. The classical PCA assumes independently and identically distributed (i.i.d.) Gaussian noise and adopts the sum of squared differences as the loss function, as shown in \refEq{eq:mafit_l2}. Since the least-squares fitting is sensitive to outliers, the classical PCA can be easily corrupted by these gross errors. For example, the reconstructed face images would include artifacts caused by the glasses or shadows in the input images \citep{de2003framework}. Recovering a subspace or low-rank matrix robustly in the presence of outliers has become a popularly-studied issue. This problem is often called {\bf Robust Principal Component Analysis} (RPCA).

\vspace{1em}
In recent years, many new techniques for low-rank matrix recovery have been proposed. In the following, we will introduce some representative works. Basically, they can be divided into two categories based on their approaches to modeling the low-rank prior. The first approach is to minimize the rank of the unknown matrix subject to some constraints. The rank minimization is often achieved by convex relaxation. We call these methods rank minimization methods. The second approach is to factorize the unknown matrix as a product of two factor matrices. The rank of the unknown matrix is upper bounded by the ranks of the factor matrices. We call these methods matrix factorization methods.

The rest of this paper is organized as follows.\footnote{A conference version of this paper appeared in Proceedings of SPIE Defense, Security, and Sensing 2013 \citep{zhou2013low}.} In Section 2, we will review the rank minimization methods for low-rank matrix recovery. We shall introduce some typical models as well as the corresponding optimization algorithms to solve these models. In Section 3, we will introduce matrix factorization methods for low-rank matrix recovery. In Section 4, we will use synthesized experiments to illustrate the performances of the discussed methods. In Section 5, we will give a brief review of the applications of low-rank modeling in image analysis. Finally, we will conclude the paper with discussions in Section 6.

\section{Rank Minimization}

A direct approach to recovering a low-rank matrix is to minimize the rank of the matrix with certain constraints that make the estimated matrix consistent with original data. However, the rank minimization problem is combinatorial and known to be NP-hard \citep{fazel2002matrix}. Therefore, convex relaxation is often used to make the minimization tractable. The most popular choice is to replace rank with the nuclear norm which is defined as
\begin{align}
\|\bfX\|_* = \sum_{i=1}^{r}\sigma_i,
\end{align}
where $\sigma_1,\sigma_2,\cdots,\sigma_r$ are the singular values of $\bfX$ and $r$ is the rank of $\bfX$. The advantages of using the nuclear norm relaxation are mainly two-folds. Firstly, the nuclear norm is convex. Hence, it is feasible to compute the global optima of the relaxed problem efficiently. Secondly, the nuclear norm is proven to be the tightest convex surrogate of rank \citep{fazel2002matrix}. It means that the nuclear norm is the best approximation to the rank operator in all convex functions. Moreover, the analogy between using the nuclear norm for low-rank matrix recovery and using the $\ell_1$-norm for sparse signal recovery has been well established \citep{recht2010guaranteed}, and the exact recovery property has been proven for some low-rank models using the nuclear norm \citep{recht2010guaranteed,candes2009exact,candes2011robust}. In the following, we will first introduce the convex models, summarize the optimization algorithms, and finally introduce some nonconvex relaxation methods briefly.

\subsection{Matrix Completion}\label{sec:mc}

In matrix completion, missing values in a matrix are estimated given observed values $\{D_{ij}|ij\in\Omega\}$, where $\Omega$ denotes the set of observed entries. As discussed previously, the common assumption is that the matrix should be low-rank. To solve the problem, the following optimization problem is often considered:
\begin{align}\label{eq:mc_rm}
\min_{\bfX}~ &\rank{\bfX}, \nonumber \\
\st & \Proj_{\Omega}(\bfX) = \Proj_{\Omega}(\bfD),
\end{align}
where $\Proj_{\Omega}(\bfX)$ denotes the operation of projecting matrix $\bfX$ to the space of all matrices with nonzero elements restricted in $\Omega$, i.e. $\Proj_{\Omega}(\bfX)$ has the same values as $\bfX$ for the entries in $\Omega$ and zero values for the entries outside $\Omega$. The equality constraint in \refEq{eq:mc_rm} says that the estimated values should coincide with the existing data.

As we discussed earlier, replacing rank with the nuclear norm can make the problem tractable. In some recent works \citep{candes2009exact,cai2010singular}, the following convex problem is solved:
\begin{align}\label{eq:mc_nm}
\min_{\bfX}~ &\|\bfX\|_*, \nonumber \\
\st & \Proj_{\Omega}(\bfX) = \Proj_{\Omega}(\bfD),
\end{align}
\hl{\citet{candes2009exact} theoretically proved that the solution of \refEq{eq:mc_nm} can exactly recover the low-rank matrix with a high probability, if the underlying low-rank matrix satisfies the incoherence condition and the locations of observed entries $\Omega$ are uniformly distributed with $|\Omega|\geq C n^{1.2}r\log{n}$ , where $|\Omega|$ is the number of observed entries, $C$ is a positive constant, $n$ is the matrix size, and $r$ is the rank. Here the incoherence condition is used to mathematically characterize the difficulty of recovering the underlying low-rank matrix from a small number of sampled entries. Informally, it says that the singular vectors of the underlying low-rank matrix should sufficiently ``spread out'' and be uncorrelated with the standard basis. An extreme example is that the underlying low-rank matrix takes 1 in its ($i,j$)-th entry and 0 elsewhere. This matrix can be recovered only if the ($i,j$)-th entry is actually sampled. This result has been further strengthened to $|\Omega|\geq C nr \mbox{ poly}(\log{n})$ by imposing the strong incoherence condition \citep{candes2010power,gross2011recovering}.}

In real applications, the observed entries may be noisy, and the equality constraint in \refEq{eq:mc_nm} will be too strict, resulting in over-fitting \citep{mazumder2010spectral}. Therefore, the following relaxed form of \refEq{eq:mc_nm} is often considered for matrix completion with noise \citep{candes2010matrix,mazumder2010spectral}:
\begin{align}\label{eq:mcn}
\min_{\bfX}~ \half\|\Proj_{\Omega}(\bfD)-\Proj_{\Omega}(\bfX)\|_F^2 + \lambda\|\bfX\|_*,
\end{align}
where the parameter $\lambda$ controls the rank of $\bfX$ and the selection of $\lambda$ should depend on the noise level \citep{candes2010matrix}.

\subsection{Robust Principal Component Analysis}

Convex programming has also been used to solve RPCA. A popular method is named sparse and low-rank decomposition \citep{candes2011robust}, and involves the decomposition of a matrix $\bfD$ as a sum of a low-rank component $\bfX$ and a sparse component $\bfE$ by minimizing the rank of $\bfX$ and the cardinality of $\bfE$ simultaneously. The surprising message is that, under some mild assumptions, the low-rank matrix can be exactly recovered by the following convex program named Principal Component Pursuit (PCP) \citep{candes2011robust}:
\begin{align}\label{eq:pcp}
\min_{\bfX,\bfE}~ &\|\bfX\|_*+\lambda\|\bfE\|_1, \nonumber \\
\st~ & \bfX+\bfE=\bfD.
\end{align}
Here, the nuclear norm $\|\bfX\|_*$ and the $\ell_1$-norm $\|\bfE\|_1$ are the convex surrogates of rank and cardinality, respectively. \citet{candes2011robust} and \citet{chandrasekaran2011rank} analyzed the conditions for exact recovery. Briefly speaking, it has been proven that the underlying low-rank matrix $\bfX^*$ and the underlying sparse matrix $\bfE^*$ can be exactly recovered with high probability if $\bfX^*$ satisfies the incoherence condition and the nonzero entries of $\bfE^*$ are sufficiently sparse with a random spatial distribution. Moreover, a theoretical choice of parameter $\lambda$ is provided to make the exact recovery most likely \citep{candes2011robust}.

The basic model in \refEq{eq:pcp} was extended to handle additional scenarios such as Stable PCP that considers Gaussian noise \citep{zhou2010stable}, the outlier pursuit that incorporates group sparsity \citep{xu2010robust}, and the matrix recovery from compressive measurements \citep{wright2012compressive}. In Stable PCP \citep{zhou2010stable}, the equality constraint in \refEq{eq:pcp} is relaxed to be $\|\bfX+\bfE-\bfD\|_F \leq \sigma$ to allow the existence of Gaussian noise. In implementation, the following problem is solved
\begin{align}\label{eq:spcp}
\min_{\bfX,\bfE}~ &\|\bfX\|_*+\lambda\|\bfE\|_1 + \frac{\mu}{2}\|\bfX+\bfE-\bfD\|_F^2,
\end{align}
where $\mu$ is a parameter determined by the noise level.

\subsection{Optimization Algorithms}

The following theorem (\citep[Theorem 2.1]{cai2010singular}) serves as an important building block in nuclear norm minimization algorithms:
\begin{theorem}\label{lemma:svt}
The solution to the following problem
\begin{align}\label{eq:prox}
    \min_{\bfX}~\half\|\bfZ-\bfX\|_F^2 + \lambda\|\bfX\|_*
\end{align}
is given by $\bfX^*=\SVT_{\lambda}(\bfZ)$, where
\begin{align}\label{eq:svt}
\SVT_{\lambda}(\bfZ) = \sum_{i=1}^{r}{(\sigma_i-\lambda)_+\bfu_i\bfv_i^T},
\end{align}
$r$ is the rank of $\bfZ$, $(x)_+ = \max(x,0)$, and $\{\bfu_i\}$, $\{\bfv_i\}$ and $\{\sigma_i\}$ are the left singular vectors, right singular vectors and singular values of $\bfZ$, respectively.
\end{theorem}
$\SVT_{\lambda}$ is named the singular value thresholding (SVT) operator \citep{cai2010singular}.

Based on \refTheor{lemma:svt}, various algorithms have been developed for specific problems. Two of the most popular techniques are the Proximal Gradient (PG) method \citep{moreau1965proximite} and the Augmented Lagrangian Method (ALM) \citep{bertsekas1999nonlinear}, both of which are applicable to a variety of convex problems. The PG method is very useful to solve the norm-regularized maximum-likelihood problems such as the model in \refEq{eq:mcn}, whose energy function comprises a differentiable loss and a nonsmooth regularizer. Moreover, the PG method is often combined with the Nesterov method to accelerate convergence \citep{nesterov2007gradient,beck2009fast}. Examples using the PG method include \citep{ji2009accelerated,mazumder2010spectral,toh2010accelerated}, etc. The ALM method is closely related to the Alternating Direction Method of Multipliers (ADMM) \citep{boyd2010distributed}. It provides a powerful framework to solve convex problems with equality constraints such as MC in \refEq{eq:mc_nm} and PCP in \refEq{eq:pcp}. The algorithms used in \citep{lin2010augmented,candes2011robust} belong to this class. The details of PG and ALM will be given in the following subsections.

\subsubsection{Proximal Gradient}

In sparse learning problems, the following optimization problem is often considered:
\begin{align}\label{eq:pg}
\min_{\bfX} f(\bfX) + \lambda\reg(\bfX),
\end{align}
where $f(\bfX)$ usually denotes a differentiable loss function and $\reg(\bfX)$ corresponds to a convex regularizer which might be nonsmooth. For example, the matrix completion with noise in \refEq{eq:mcn} uses $f(\bfX)=\|\Proj_{\Omega}(\bfX-\bfD)\|_F^2$ and $\reg(\bfX)=\|\bfX\|_*$.

If $f(\bfX)$ is simply the sum of squared differences between $\bfX$ and a given matrix, the problem in \refEq{eq:pg} is named the proximal problem \citep{moreau1965proximite}, which can be solved analytically for many types of $\reg(\bfX)$. For example, if $\reg(\bfX)$ is the nuclear norm, the problem can be solved analytically based on \refTheor{lemma:svt}.

When \refEq{eq:pg} is not a standard proximal problem, the Proximal Gradient (PG) method \citep{moreau1965proximite} is usually used. In PG, a quadratic approximation to $f(\bfX)$ is made around the previous estimate $\bfX'$ in each iteration. Define
\begin{align}
Q_{\mu}(\bfX,\bfX') &= f(\bfX') + \langle \nabla f(\bfX'), \bfX-\bfX' \rangle + \frac{\mu}{2}\|\bfX-\bfX'\|_F^2 + \lambda\reg(\bfX), \nonumber \\
&= \frac{\mu}{2}\|\bfX-[\bfX'-\frac{1}{\mu}\nabla f(\bfX')]\|_F^2 + \lambda\reg(\bfX) + const.,
\end{align}
where $<>$ means the inner product and $\mu$ is a constant. It can be proven that \citep{beck2009fast}, if $f(\bfX)$ is differentiable and convex with a Lipschitz continuous gradient satisfying
\begin{align}
    \|\nabla f(\bfX_1) - \nabla f(\bfX_2)\|_F \leq \mu \|\bfX_1 - \bfX_2\|_F,
\end{align}
\refEq{eq:pg} can be solved by repeatedly updating $\bfX$ via
\begin{align}\label{eq:ist}
    \bfX^{k+1}&=\arg\min_{\bfX}~Q_\mu(\bfX,\bfX^{k}), \nonumber \\
    & = \arg\min_{\bfX}~\frac{1}{2}\|\bfX-[\bfX^k-\frac{1}{\mu}\nabla f(\bfX^k)]\|_F^2 + \frac{\lambda}{\mu}\reg(\bfX)
\end{align}
with a convergence rate of $\mathcal{O}({1}/{k})$. It is easy to see that \refEq{eq:ist} is simply the proximal problem, which is often convenient to solve.

The Accelerated Proximal Gradient (APG) method uses the Nesterov method \citep{nesterov1983method} to accelerate the convergence of PG. Instead of making quadratic approximation around $\bfX^{k}$, APG makes the approximation at another point $\bfY^k$, which is a linear combination of $\bfX^{k}$ and $\bfX^{k-1}$. This modification will give a convergence rate of $\mathcal{O}(\frac{1}{k^2})$. Please refer to \citep{nesterov2007gradient,beck2009fast} for details. The APG method is summarized in \refAlg{alg:apg}.

\begin{algorithm}
    \caption{Accelerated Proximal Gradient (APG)}\label{alg:apg}
    \begin{algorithmic}[1]
    \algsetup{linenodelimiter=.}
        \STATE {\bf Initialize:} $\bfX^0=\bfX^{-1}\in\RR{m}{n}$, $t_0=t_{-1}=1$
        \REPEAT
            \STATE $\bfY^k = \bfX^k + \frac{t^{k-1}-1}{t^k} (\bfX^k - \bfX^{k-1})$
            \STATE $\bfX^{k+1} = \arg\min_{\bfX}Q_\mu(\bfX,\bfY^{k})$
            \STATE $t^{k+1} = \frac{1+\sqrt{1+4(t^k)^2}}{2}$
        \UNTIL{convergence}
    \end{algorithmic}
\end{algorithm}

The PG and APG methods have been intensively used to solve the matrix completion problem in \refEq{eq:mcn}, where the updating in \refEq{eq:ist} is solved via SVT. For example, the SOFT-IMPUTE algorithm in \citep{mazumder2010spectral} solves \refEq{eq:mcn} by iteratively updating $\bfX$ by:
\begin{align}\label{eq:softimpute}
    \bfX^{k+1} = \SVT_{\lambda}(\Proj_{\Omega}(\bfD)+\Proj_{\Omega^{\perp}}(\bfX^k)) = \SVT_{\lambda}(\bfX^k-[\Proj_{\Omega}(\bfX^k)-\Proj_{\Omega}(\bfD)]),
\end{align}
where $\Proj_{\Omega^{\perp}}$ denotes the complementary projection such that $\Proj_{\Omega}(\bfX^k)+\Proj_{\Omega^{\perp}}(\bfX^k)=\bfX^k$. It is straightforward to find that \refEq{eq:softimpute} is equivalent to \refEq{eq:ist} with $\mu=1$. Hence, the SOFT-IMPUTE algorithm can be interpreted as PG with fixed step length. The FPCA algorithm introduced in \citep{ma2011fixed} is also based on PG with a continuation technique to accelerate convergence. \citet{ji2009accelerated} and \citet{toh2010accelerated} also proposed different implementations of APG for matrix completion. \citet{tomioka2010fast} proposed a Dual Augmented Lagrangian algorithm for matrix completion, which achieves super-linear convergence. It can be interpreted as a proximal method with the descending directions computed from the augmented Lagrangian of the dual problem.

\subsubsection{Augmented Lagrangian Method}

The Augmented Lagrangian Method (ALM) \citep{bertsekas1999nonlinear} is a classical tool to minimize a convex function with equality constraints. We will use PCP in \refEq{eq:pcp} as an example to introduce this method.

To remove the constraint $\bfX+\bfE=\bfD$, a multiplier $\bfY$ is introduced and the augmented Lagrangian of \refEq{eq:pcp} reads
\begin{align}\label{eq:alm_pcp}
    L_{\mu}(\bfX,\bfE,\bfY) = \|\bfX\|_* + \lambda\|\bfE\|_1 + <\bfY,\bfD-\bfX-\bfE> + \frac{\mu}{2}\|\bfD-\bfX-\bfE\|_F^2.
\end{align}
ALM alternates between the following two steps:
\begin{align}
    &(\bfX^{k+1},\bfE^{k+1}) = \arg\min_{\bfX,\bfE} L_{\mu}(\bfX,\bfE,\bfY^{k}), \label{eq:alm_p} \\
    &\bfY^{k+1} = \bfY^{k} + \mu(\bfD-\bfX^{k+1}-\bfE^{k+1}), \label{eq:alm_d}
\end{align}
which are named primal minimization and dual ascent, respectively. For PCP, the primal minimization in \refEq{eq:alm_p} is difficult over $\bfX$ and $\bfE$ simultaneously. But if we fix one variable and minimize over the other one, the marginal optimization over $\bfX$ (or $\bfE$) turns into the nuclear norm (or $\ell_1$-norm) regularized proximal problem, which can be efficiently solved by SVT (or soft-thresholding). Then, the $\bfX$-step and $\bfE$-step are repeated until convergence to solve \refEq{eq:alm_p}.

A more efficient way is to update the primal variables $\bfX$ and $\bfE$ for only one iteration, instead of exactly solving \refEq{eq:alm_p} before updating the dual variable $\bfY$  \citep{lin2010augmented,candes2011robust}. This is named the Inexact Augmented Lagrangian Method (IALM), a special case of the Alternating Direction Method of Multipliers (ADMM) \citep{boyd2010distributed}. The method is summarized in \refAlg{alg:admm_pcp}. It can be proven that the sequences $\{\bfX^{k}\}$ and $\{\bfE^{k}\}$ will converge to an optimal solution of \refEq{eq:pcp} \citep{lin2010augmented,boyd2010distributed}.

\begin{algorithm}
    \caption{Inexact Augmented Lagrangian Method (IALM) for PCP}\label{alg:admm_pcp}
    \begin{algorithmic}[1]
    \algsetup{linenodelimiter=.}
        \STATE {\bf Initialize:} $\bfE^0=\bfY^0=\mathbf{0}$
        \REPEAT
            \STATE $\bfX^{k+1} = \arg\min_{\bfX} L_{\mu}(\bfX,\bfE^k,\bfY^{k})$
            \STATE $\bfE^{k+1} = \arg\min_{\bfE} L_{\mu}(\bfX^{k+1},\bfE,\bfY^{k})$
            \STATE $\bfY^{k+1} = \bfY^{k} + \mu(\bfD-\bfX^{k+1}-\bfE^{k+1})$
        \UNTIL{convergence}
    \end{algorithmic}
\end{algorithm}

ALM can also be applied to matrix completion in \refEq{eq:mc_nm}. In \citep{lin2010augmented}, the equality constraint $\Proj_{\Omega}(\bfX) = \Proj_{\Omega}(\bfD)$ is replaced by $\bfX = \bfD+\bfE$ and $\Proj_{\Omega}(\bfE)=\mathbf{0}$. The new constraint is equivalent to the original one, but the projection operator on $\bfX$ has been removed. Then, the ALM is applied. In this way, minimizing the augmented Lagrangian over $\bfX$ turns into a proximal problem, which could be solved by SVT. ALM was also applied to solve the nonnegative matrix factorization problem for matrix completion \citep{xu2012alternating}.

\subsection{Nonconvex Rank Minimization}

Recently, a few works used nonconvex functions instead of the nuclear norm as the surrogates of rank for rank minimization. A typical choice is the Schatten-p norm of singular values:
\begin{align}
f_p(\bfX) = \left(\sum_{i=1}^{r}\sigma_i^p\right)^{1/p},
\end{align}
where $\sigma_1,\cdots,\sigma_r$ are singular values of $\bfX$. When $p\rightarrow 0$, $f_p(\bfX)\rightarrow \rank{\bfX}$, and the minimization is intractable. When $p=1$, $f_1(\bfX)$ turns out to be the nuclear norm, which is the tightest convex surrogate. To bridge the gap, the nonconvex cases with $0<p<1$ were considered in recent literature \citep{majumdar2011algorithm,mohan2012iterative,marjanovic2012on,nie2012robust,li2014reweighted}. The theoretical analysis on the recovery properties can be found in \citep{zhang2013restricted}. Besides the Schatten-p, other nonconvex surrogate functions for rank minimization were also studied in \citep{lu2014generalized}.

\section{Matrix Factorization}

Instead of minimizing rank, another approach to modeling the low-rank property is matrix factorization. Matrix factorization intends to decompose $\bfX\in\RR{m}{n}$ as a product of two factor matrices $\bfX=\bfA\bfB^T$, where $\bfA\in\RR{m}{r}$ and $\bfB\in\RR{n}{r}$. Using matrix factorization to model a low-rank matrix is based on the fact that
\begin{align}
\rank{\bfA\bfB^T} \leq \min\left(\rank{\bfA},\rank{\bfB}\right).
\end{align}
Therefore, if $r$ is small, $\bfX$ has a small rank. Finally, the problem of recovering a low-rank matrix can be converted into estimating two factor matrices $\bfA$ and $\bfB$. In this paper, we will discuss representative matrix-factorization methods in the context of low-rank matrix recovery. Notice that not all matrix factorization methods aim to recover a low-rank matrix. For example, the outputs of nonnegative matrix factorization \citep{lee1999learning} or dictionary learning \cite{tosic2011dictionary} are not necessarily low-rank. We will not discuss these methods here. For a summary of matrix factorization methods, please refer to \citep{singh2008unified}.

\subsection{Matrix Factorization with Missing Values}

Basically, the factorization-based methods for matrix completion aim to solve the following optimization problem:

\begin{align}\label{eq:mc_mf}
\min_{\bfA,\bfB} \half\|\Proj_{\Omega}(\bfD)-\Proj_{\Omega}(\bfA\bfB^T)\|_F^2.
\end{align}

A straightforward approach to solving \refEq{eq:mc_mf} is to minimize the function over $\bfA$ or $\bfB$ alternately by fixing the other one. Each subproblem of estimating $\bfA$ or $\bfB$ turns into a least-squares problem which admits a closed-form solution. Algorithms of this type have been extensively studied in many works such as the early computer vision literature \citep{shum1995principal,vidal2004motion} and the recent matrix recovery literature \citep{haldar2009rank,tang2011lower,jain2013low}.

The matrix completion solver LMaFit \citep{wen2012solving} also adopted the alternating strategy to solve the following equivalent form of \refEq{eq:mc_mf}:
\begin{align}
    &\min_{\bfA,\bfB, \bfZ} \half\|\bfZ-\bfA\bfB^T\|_F^2, \nonumber \\
    &\st~ \Proj_{\Omega}(\bfZ) = \Proj_{\Omega}(\bfD),
\end{align}
where $\bfZ$ is an auxiliary variable. Each step of updating $\bfA$, $\bfB$ or $\bfZ$ can be solved very efficiently. Additionally, LMaFit integrates a nonlinear successive over-relaxation scheme to accelerate the convergence of alternation.

\hl{While the formulation in \refEq{eq:mc_mf} is nonconvex, the empirical results in many works demonstrated that the alternating minimization performed both accurately and efficiently compared to convex methods \citep{haldar2009rank,keshavan2009low,tang2011lower}. Meanwhile, the theoretical analysis in \citep{jain2013low} showed that the alternating minimization can succeed under the conditions similar to the existing conditions given in \citep{candes2009exact}, which has been introduced in \refSec{sec:mc}. The lower bounds for the recovery error of using alternating minimization for matrix completion were analyzed in \citep{tang2011lower}.}

In computer vision literature, many works adopted higher order algorithms instead of alternating least squares to solve \refEq{eq:mc_mf} for faster convergence and better precision. For example, \citet{buchanan2005damped} developed a Damped Newton algorithm to solve the problem. The variables $\bfA$ and $\bfB$ are updated based on the Newton algorithm with a damping factor. However, they cannot handle large-scale problems due to the infeasibility of computing the Hessian matrix over a large number of variables. To interpolate between the alternating least squares and the Newton algorithm, some works proposed to use hybrid algorithms. In the Wiberg algorithm \citep{okatani2007wiberg}, $\bfA$ is updated via least squares while $\bfB$ is updated by a Gauss-Newton step in each iteration. Later, the Wiberg algorithm was extended to a damped version to achieve better convergence \citep{okatani2011efficient}. \citet{chen2008optimization} proposed an algorithm similar to the Wiberg algorithm. The difference is that $\bfB$ is updated via the Levenberg-Marquadt algorithm and constrained in $\{\bfB|\bfB^T\bfB=\bfI\}$. Interested readers can refer to \citep{okatani2011efficient} for a more detailed introduction to the factorization methods in computer vision.

When observation is highly incomplete, the problem in \refEq{eq:mc_mf} is likely to be ill-posed, which is a common case in collaborative filtering. A popular approach to addressing this issue is to penalize the squared Frobenious norms of factor matrices:
\begin{align}\label{eq:mmmf}
\min_{\bfA,\bfB} \half\|\Proj_{\Omega}(\bfD)-\Proj_{\Omega}(\bfA \bfB^T)\|_F^2 + \frac{\lambda}{2}(\|\bfA\|_F^2+\|\bfB\|_F^2).
\end{align}
This method is named Maximum Margin Matrix Factorization (MMMF) \citep{srebro2005maximum}. The idea is similar to using the squared $\ell_2$-norm in ridge regression to improve the stability of parameter estimation. Moreover, the following equality is established in \citep{srebro2005maximum}
\begin{align}\label{nuclear_fro}
\|\bfX\|_* = \min_{\bfA,\bfB:\bfX=\bfA \bfB^T}\half(\|\bfA\|_F^2+\|\bfB\|_F^2),
\end{align}
which indicates the equivalence between the MMMF in \refEq{eq:mmmf} and the nuclear norm minimization in \refEq{eq:mcn}. This equivalence was also studied in \citep{mazumder2010spectral,wang2012probabilistic,cabral2013unifying}. To solve the optimization in \refEq{eq:mmmf}, either gradient-based algorithms \citep{rennie2005fast} or alternating minimization can be used \citep{wang2012probabilistic,cabral2013unifying}.

\subsection{Riemannian Optimization}\label{sec:manifold}

Another widely-used regularization strategy in low-rank matrix factorization is to constrain the search space and optimize over manifolds.

\citet{keshavan2010matrix} proposed to solve the following matrix completion problem
\begin{align}\label{eq:optspace}
\min_{\bfA,\Sigma,\bfB} &\half\|\Proj_{\Omega}(\bfD)-\Proj_{\Omega}(\bfA\Sigma\bfB^T)\|_F^2, \nonumber \\
\st~ &\bfA\in\mbox{Gr}(r,m),~\bfB\in\mbox{Gr}(r,n),~\Sigma\in\RR{r}{r},
\end{align}
where $\mbox{Gr}(r,n)$ denotes the set of $r$-dimensional subspaces in $\R{n}$, which forms a Riemannian manifold named Grassmannian. \citet{keshavan2010matrix} proposed an algorithm named OptSpace to iteratively estimate the factor matrices, where $\bfA$ and $\bfB$ are updated by gradient descent over the Grassmannian, while $\Sigma$ is updated by least squares. Similar to the theoretical results for the convex method \citep{candes2010matrix}, \citet{keshavan2010matrixa} provided the performance guarantee of OptSpace under an appropriate incoherence condition. Building upon the same model, \citet{ngo2012scaled} proposed a scaled-gradient procedure to accelerate the convergence of the algorithm. \citet{dai2011subspace} proposed an algorithm named SET to solve the following two-factor model:
\begin{align}\label{eq:set}
\min_{\bfA,\bfB} &\half\|\Proj_{\Omega}(\bfD)-\Proj_{\Omega}(\bfA\bfB^T)\|_F^2, \nonumber \\
\st~ &\bfA\in\mbox{Gr}(r,m),~\bfB\in\RR{n}{r}.
\end{align}
SET updates $\bfA$ over the Grassmannian and estimates $\bfB$ by least squares. Based on the same model, \citet{boumal2011rtrmc} developed an algorithm named RTRMC that optimizes the cost function by a second-order Riemannian trust-region method to achieve faster convergence.

\citet{mishra2013fixed} proposed a framework of optimization over Riemannian quotient manifolds for low-rank matrix factorization. They investigated three types of matrix factorization: the full-rank factorization, the polar factorization and the subspace-projection factorization, which are related to the models in \refEq{eq:mc_mf}, \refEq{eq:optspace} and \refEq{eq:set}, respectively. To take into account the invariance over a class of equivalent solutions, they explored the underlying quotient nature of the search spaces and designed a class of gradient-based and trust-region algorithms over the quotient search spaces. They concluded through experiments that the three factorization models with different Riemmnanian structures were almost equivalent in terms of computational complexity and performed favorably compared to the previous methods such as LMaFit and OptSpace. More related works include \citep{meyer2011linear,mishra2012riemannian,mishra2013low,absil2013two}, etc.

Instead of exploring the geometries of search spaces of factor matrices, \citet{vandereycken2013low} and \citet{shalit2012online} proposed to directly optimize a function over the set of fixed-rank matrices:
\begin{align}
\min_{\bfX} ~ & f(\bfX) = \half \|\Proj_{\Omega}(\bfD)-\Proj_{\Omega}(\bfX)\|_2^2, \nonumber \\
\st ~ & \bfX \in \mathcal{M}_r.
\end{align}
Here $\mathcal{M}_r$ denotes the set of rank-$r$ matrices in $\RR{m}{n}$, which forms a smooth manifold. \citet{vandereycken2013low} developed a conjugate gradient descent algorithm named LRGeomCG to efficiently optimize any smooth function over $\mathcal{M}_r$, while \citet{shalit2012online} designed an online algorithm to solve large-scale problems. Numerical experiments \citep{vandereycken2013low,mishra2013fixed} showed that LRGeomCG performed comparably with the quotient-space methods for matrix completion.

Recently, a manifold-optimization toolbox named Manopt has been developed \citep{boumal2014manopt}, providing a lot of ready-to-use algorithms to solve optimization problems over various manifolds, such as Grassmannians and the fixed-rank manifolds.

\subsection{Robust Matrix Factorization}

Robust matrix factorization is a method for handling outliers in data, and can be regarded as the factorization approach towards RPCA. As mentioned before, the sensitivity to outliers for traditional methods is due to the squared loss used in \refEq{eq:mc_mf}, which penalizes large errors too much, resulting in biased fitting. To address this issue, a typical approach is to use more robust loss functions:
\begin{align}\label{eq:rmf}
\min_{\bfA,\bfB}~ \sum_{ij} \rho(D_{ij}-[\bfA\bfB^T]_{ij}),
\end{align}
where $[\bfA\bfB^T]_{ij}$ denotes the entry $ij$ of $\bfA\bfB^T$ and $\rho$ is a robust loss function. For example, the Geman-McClure function defined as $\rho(x)=\frac{x^2}{2(1+x^2)}$ is adopted in \citep{de2003framework}. To solve the optimization problem, alternating minimization is carried out, where $\bfA$ and $\bfB$ are updated iteratively by solving robust linear regression via iterative reweighted least squares. A similar idea of reweighting data based on robust estimators is used in \citep{aanaes2002robust}. \citet{ke2005robust} adopted the $\ell_1$-penalty and solved the problem by alternating $\ell_1$-minimization.

\hl{More examples of using the $\ell_1$-norm for robust matrix factorization include \citep{kwak2008principal,eriksson2010efficient,zheng2012practical,shu2014robust}. \citet{kwak2008principal} proposed to maximize the $\ell_1$-norm of the projection of a data point onto the unknown principal directions instead of minimizing the residue in \refEq{eq:rmf}. \citet{eriksson2010efficient} generalized the Wiberg algorithm \citep{okatani2007wiberg} to handle the $\ell_1$ case. \citet{zheng2012practical} proposed to add a nuclear-norm regularizer to improve the convergence and solved the optimization by ALM. \citet{shu2014robust} presented an efficient algorithm using the group-sparsity regularization and established the equivalence between the proposed method and rank minimization. Many other works tried to address the problem from the probabilistic point of view, which modeled non-Gaussian noise to improve robustness. We will discuss them in \refSec{sec:pmf}.}

\subsection{Online and Parallel Algorithms}

The demand of online processing of streaming data like a video has motivated the development of many incremental algorithms for subspace tracking in past decades \citep{bunch1978updating,comon1990tracking,edelman1998geometry}. In more recent works such as \citep{brand2002incremental,balzano2010online,he2012incremental}, online algorithms for incomplete or corrupted data were developed. In GROUSE \citep{balzano2010online} for example, the following cost function is minimized over a subspace $\bfA$ each time a new data point $\bfd_t$ arrives:
\begin{align}\label{eq:imf}
F^{(t)}(\bfA) = \min_{\bfb_t} ~ \|\Proj_{\Omega_t}(\bfd_t-\bfA\bfb_t)\|_2^2,
\end{align}
where $\Omega_t$ denotes the set of observed entries in $\bfd_t$. In each iteration, $\bfb_t$ is first solved via least squares, and $\bfA$ is then updated via gradient descent over the Grassmannian. The algorithm GRASTA introduced in \citep{he2012incremental} extends GROUSE to handle outliers for robust estimation by replacing the squared loss with the $\ell_1$-norm in \refEq{eq:imf}. To solve the $\ell_1$-minimization problem in each iteration, the ADMM framework is used in GRASTA. \citet{shalit2012online} developed an online algorithm by optimization over the low-rank manifold, which has been discussed in \refSec{sec:manifold}.

To improve the scalability of matrix factorization algorithms, some parallel frameworks have been proposed, which can perform matrix factorization in parallel computing architectures to handle extremely large datasets. \citet{recht2013parallel} proposed an algorithm named JELLYFISH for large-scale matrix completion. It minimizes the energy function in \refEq{eq:mmmf} via incremental gradient descent, i.e. the variables are updated following an approximate gradient constructed from a sampling of matrix entries. Moreover, JELLYFISH adopts a block matrix-partitioning scheme with a special sampling order to allow a parallel implementation of the algorithm on multiple cores to achieve a speed-up nearly proportional to the number of cores. \citet{gemulla2011large} adopted similar strategies for stochastic and parallel implementation. \citet{mackey2011divide} introduced a divide-and-conquer framework for matrix factorization. The framework first divides a large input matrix into smaller submatrices (e.g. by selecting columns and rows), factorizes submatrices in parallel using existing matrix factorization algorithms, and finally combines the solutions of the subproblems using random matrix approximation techniques. They provided a theoretical analysis on the recovery probability of the paralleled algorithm compared to its batch version.

\subsection{Probabilistic Matrix Factorization}\label{sec:pmf}

In this subsection, we shall briefly introduce a class of methods that treat low-rank matrix factorization from a probabilistic point of view.

\subsubsection{Probabilistic PCA}\label{sect:prob-pca}

Probabilistic PCA (PPCA) \citep{tipping1999probabilistic} is a latent variable model which successfully formulates classical PCA \citep{jolliffe2002principal,hotelling1933analysis} into the probabilistic framework. Let $\bfd_i\in \R{m}$ be the $i$-th observed data point and $\bfb_i\in \R{r}$ be the $i$-th latent variable in the latent space. PPCA assumes that $\bfd_i$ is linearly related to $\bfb_i$ by a matrix $\bfA \in \R{m\times r}$:
\begin{equation}\label{LVM-linear2}
  \bfd_i = \bfA \bfb_i + \bfe_i,
\end{equation}
where $\bfe_i\in \R{m}$ denotes random noise, which follows the Gaussian distribution $\mathcal{N}(\bf0, \beta^{-1}\bfI)$ with $\beta$ denoting the precision. Then, the likelihood of model (\ref{LVM-linear2}) can be written as
\begin{equation}\label{}
  \Pr(\bfd_i|\bfA, \bfb_i, \beta) = \mathcal{N}(\bfd_i| \bfA\bfb_i, \beta^{-1}\bfI).
\end{equation}
To obtain the marginalized likelihood of $\bfA$, we need to integrate out $\bfb_i$, and a prior distribution of $\bfb_i$ has to be specified. PPCA adopts a zero mean and unit covariance Gaussian distribution as the prior. After some derivation, the marginal likelihood reads:
\begin{align}\label{LVM-linear3}
  \Pr(\bfd_i|\bfA, \beta) &= \int{\Pr(\bfd_i|\bfA, \bfb_i, \beta)\Pr(\bfb_i)}d\bfb_i, \nonumber \\
  &= \mathcal{N}(\bf0, \bfA\bfA^T+\beta^{-1}\bfI).
\end{align}
By further assuming the independence of data points, the likelihood of full data is given by
\begin{equation}\label{LVM-linear4}
  \Pr(\bfD|\bfA, \beta) = \prod^{n}_{i=1}\Pr(\bfd_i|\bfA, \beta).
\end{equation}
Finally, the matrix $\bfA$ can be estimated by maximizing the above likelihood function. The maximum likelihood estimate (MLE) can be obtained analytically as
\begin{equation}\label{}
  \hat{\bfA} = \bfU_r(\Sigma_r -\beta^{-1} \bfI)^{1/2}\bfQ,
\end{equation}
where the columns of $\bfU_r\in \R{m\times r}$ are given by the $r$ eigenvectors of the covariance matrix $\bfD^T\bfD$ corresponding to the $r$ largest eigenvalues $\lambda_1,\cdots,\lambda_r$, $\Sigma_r$ is a diagonal matrix with the diagonal elements being $\lambda_1,\cdots,\lambda_r$, and $\bfQ$ is an arbitrary $r\times r$ orthogonal matrix. As a result, the column space of $\hat\bfA$ is identical to the subspace spanned by the principal components derived from the classical PCA. The MLE of the noise variance is given by
\begin{equation}\label{}
  \beta^{-1} = \frac{1}{n-r}\sum^n_{i=r+1}{\lambda_i},
\end{equation}
which can be interpreted as the average variance of the rest dimensions.

An advantage of the probabilistic formulation of PCA is that it can be very helpful to automatically choose $r$ by seeking a Bayesian approach \citep{bishop1999bayesian}. Consider an independent Gaussian prior over each column of $\bfA$:
\begin{equation}\label{}
  \Pr(\bfA|\bfgamma)=\prod^r_{i=1}{\left(\frac{\gamma_i}{2\pi}\right)^{m/2}}\exp\left(-\frac{1}{2}\gamma_i\bfa^T_i\bfa_i\right).
\end{equation}
where $\bfa_i\in \R{m}$ is the $i$-th column of $\bfA$. Notice that the hyper-parameter $\gamma_i$ controls the inverse variance of $\bfa_i$. When $\gamma_i$ converges to a large value during the inference, it implies that the variance of $\bfa_i$ is small, and consequently the dimension spanned by $\bfa_i$ is automatically switched off. An empirical Bayesian approach can be used to find $\bfgamma$, which is also known as automatic relevance determination (ARD) in the machine learning literature \citep{mackay1995probable,bishop2006pattern}. The well-known sparse learning model, relevance vector machines (RVM) \citep{tipping2001sparse}, also adopts a similar technique to optimize the hyperparameters.

The estimation of latent variables $\bfb_i$ can be treated similarly by integrating out $\bfA$, which is known as dual PPCA \citep{lawrence2005probabilistic}.

\subsubsection{Probabilistic Matrix Completion and RPCA}

Probabilistic matrix factorization methods generally consider the following model to describe the observed data:
\begin{align}
\bfD=\bfA\bfB^T+\bfE.
\end{align}

The probabilistic framework can naturally handle missing values in matrix completion by only considering the likelihood over the observed entries
\begin{align}
\Pr(\bfD|\Theta) = \prod_{ij\in\Omega}\Pr(D_{ij}|\Theta),
\end{align}
where $\Theta$ is the set of parameters and $\Omega$ is the set of observed entries.

A representative work is the Probabilistic Matrix Factorization (PMF) \citep{salakhutdinov2008probabilistic}. It assumes the following priors:
\begin{align}\label{eq:pmf_model}
A_{ij} &\sim \mathcal{N}(0,\gamma_a^{-1}),\nonumber \\
B_{ij} &\sim \mathcal{N}(0,\gamma_b^{-1}),\nonumber \\
E_{ij} &\sim \mathcal{N}(0,\beta^{-1}),
\end{align}
where $\gamma_a$, $\gamma_b$ and $\beta$ are hyperparameters. If treating the hyperparameters as fixed values, we have the following posterior probability of $\bfA$ and $\bfB$
\begin{align}\label{eq:pmf_posterior}
  \Pr(\bfA,\bfB &| \bfD,\gamma_a,\gamma_b,\beta) \propto \Pr(\bfD|\bfA,\bfB,\beta)\Pr(\bfA|\gamma_a)\Pr(\bfB|\gamma_b) \nonumber \\
  &= \prod_{ij\in\Omega}\mathcal{N}(D_{ij}|[\bfA\bfB^T]_{ij},\beta^{-1})
  \prod_{ij}\mathcal{N}(A_{ij}|0,\gamma_a^{-1})\prod_{ij}\mathcal{N}(B_{ij}|0,\gamma_b^{-1}).
\end{align}
After simple derivation, we can see that the maximum a posteriori (MAP) estimate of model \refEq{eq:pmf_posterior} turns out to be the solution of \refEq{eq:mmmf} with $\lambda = \gamma_a/\beta = \gamma_b/\beta$. This gives a probabilistic interpretation to MMMF. The regularization in MMMF corresponds to imposing Gaussian priors on $\bfA$ and $\bfB$.

The advantage of probabilistic modeling is that the regularization parameters $\gamma_a$, $\gamma_b$ and $\beta$ do not need to be predefined. They can be automatically determined by treating them as variables, introducing priors on them and estimating them from the data \citep{salakhutdinov2008probabilistic}. Later, \citet{salakhutdinov2008bayesian} proposed a full Bayesian method for PMF and solved the model by Markov Chain Monte Carlo (MCMC) sampling.

For RPCA, the prior on $\bfE$ needs to be changed. For instance, \citet{wang2012probabilistic} proposed Probabilistic Robust Matrix Factorization (PRMF) that used the Laplacian prior to model the error
\begin{equation}
 \Pr(E_{ij}|\beta)=\left(\frac{\beta}{2}\right)\exp(-\beta |E_{ij}|).
\end{equation}
Compared to the Gaussian prior, the Laplacian prior will encourage $\bfE$ to be sparse and allow $E_{ij}$ to have a large magnitude. The MAP estimate is given by
\begin{equation}\label{eq:rpmf}
  \min_{\bfA,\bfB} \|\bfD- \bfA\bfB^T\|_1 + \lambda\|\bfA\|^2_F + \lambda \|\bfB\|^2_F.
\end{equation}
Notice that the Laplacian prior on the error term induces the $\ell_1$-penalty on the residue. One can see the connection between PRMF and PCP by \refEq{nuclear_fro}.

\citet{babacan2012sparse} proposed a Bayesian method to solve RPCA. It assumes the model $\bfD=\bfA\bfB^T + \bfE + \bfZ$, where $\bfE$ is a sparse matrix used to model outliers and $\bfZ$ is a dense matrix used to model noise. Suppose $Z_{ij}$ is i.i.d. Gaussian noise following $\mathcal{N}(0,\beta^{-1})$ , the conditional probability of the observation is given by
\begin{equation}\label{}
  \Pr(\bfD|\bfA,\bfB,\bfE,\beta) = \mathcal{N}(\bfD|\bfA\bfB^T+\bfE, \beta)\propto \exp\left(-\frac{\beta}{2}\|\bfD-\bfA\bfB^T-\bfE\|_F^2\right).
\end{equation}
Moreover, the following priors are assumed
\begin{align}\label{eq:vbmf}
  \Pr(\bfE|\bfalpha)&= \prod_{ij}\mathcal{N}(E_{ij}|0,\alpha^{-1}_{ij}), \nonumber \\
  \Pr(\bfA|\bfgamma)&=\prod_i\mathcal{N}(\bfa_{i}|\mathbf{0},\gamma^{-1}_i \bfI), \nonumber\\
  \Pr(\bfB|\bfgamma)&=\prod_i\mathcal{N}(\bfb_{i}|\mathbf{0},\gamma^{-1}_i \bfI).
\end{align}
Instead of using fixed values, the hyperparameters $\alpha_{ij}$ and $\gamma_i$ are further modeled using Gamma priors. Finally, a variational algorithm is used to estimate $\bfA$ and $\bfB$ as well as the hyperparameters.

Other probabilistic methods for robust matrix factorization include \citep{lakshminarayanan2011robust,ding2011bayesian,wang2013bayesian,meng2013robust}, etc.

In probabilistic matrix factorization, the number of columns of $\bfA$ and $\bfB$ does not necessarily determine the rank of $\bfA\bfB^T$. It only serves as an upper bound. During the inference, the true rank will be determined automatically \citep{babacan2012sparse}. The hierarchical modeling with hyperparameters plays an important role in the automatic determination of rank. During the inference, some $\gamma_i$ in \refEq{eq:vbmf} will converge to extremely large values, resulting in the corresponding columns being close to zero. The automatic ``switch-off" of these columns driven by the data will determine the final rank of $\bfA\bfB^T$.

\subsection{Projection-Based Methods}

Another category of methods estimate a low-rank matrix under an explicit rank constraint. To make the constraint satisfied during optimization, these methods often use a greedy strategy that projects intermediate results to the feasible set of the rank constraint. While conceptually these methods use an explicit rank constraint, numerically they implement the low-rank projection step using factorization methods.

In \citep{jain2010guaranteed}, the following problem is considered:
\begin{align}\label{eq:lra}
\min_{\bfX} ~ & f(\bfX) = \half \|\mathcal{A}(\bfX)-\bfb\|_2^2, \nonumber \\
\st ~ & \rank{\bfX} \leq r,
\end{align}
where $\mathcal{A}$ is a linear operator on $\bfX$. Notice that the matrix completion problem is a special case of \refEq{eq:lra}. An algorithm named Singular Value Projection (SVP) was proposed in \citep{jain2010guaranteed} to solve \refEq{eq:lra}. It uses a projected gradient descent scheme which alternates between updating $\bfX$ via gradient descent and projecting the intermediate result to the set of rank-$r$ matrices. By the matrix approximation theorem in \refEq{eq:mal}, the projection is done by calculating the SVD of the matrix and keeping the $r$ largest singular values. This procedure is similar to the proximal gradient method for nuclear norm minimization by replacing SVT with the low-rank projection. The soft thresholding of singular values is applied in SVT, while the hard thresholding of singular values is used in SVP. Therefore, SVP is analogous to the iterative hard thresholding algorithm in sparse coding \citep{blumensath2009iterative}.

The ADMiRA algorithm introduced in \citep{lee2010admira} also intends to solve the problem in \refEq{eq:lra}. Instead of using hard thresholding as in SVP, ADMiRA extends the CoSaMP algorithm \citep{needell2009cosamp} in compressive sensing to the matrix case. The matching pursuit-like \citep{mallat1993matching} scheme is used to stepwise select the basis vectors to reconstruct the column space of $\bfX$, which can minimize the function in \refEq{eq:lra}. The SpaRCS algorithm proposed in \citep{waters2011sparcs} can be regarded as a counterpart of ADMiRA to solve the robust matrix factorization problem.

The GoDec algorithm proposed in \citep{zhou2011godec} uses iterative hard thresholding to solve the following nonconvex formulation of RPCA:
\begin{align}\label{eq:godec}
    \min_{\bfX,\bfE}~ &\|\bfD-\bfX-\bfE\|_F^2 \nonumber \\
    \st~ & \rank{\bfX} \leq r, \nonumber \\
    & \|\bfE\|_0 \leq k.
\end{align}
To minimize \refEq{eq:godec}, GoDec alternates between the low-rank projection to estimate $\bfX$ and the hard thresholding to estimate $\bfE$. To avoid the computation of SVD, GoDec uses a bilateral random projection scheme to compute the low-rank projection.

\begin{table}
\renewcommand{\arraystretch}{1.25}
\tbl{Representative Low-Rank Matrix Recovery Algorithms.\label{tab:alg}}{
\begin{tabular}{p{1.6cm}llp{4cm}}
\toprule
Category & Algorithm $\&$ reference & Problem & Main techniques \\
\midrule
Rank & SVT \citep{cai2010singular} & MC & Proximal gradient (PG) \\
Minimization & FPCA \citep{ma2011fixed} & MC & PG, approximate SVD \\
& SOFT-IMPUTE \citep{mazumder2010spectral} & MC & PG, warm-start\\
& APG \citep{ji2009accelerated,toh2010accelerated} & MC & Accelerated PG \\
& PCP \citep{candes2011robust} & RPCA & Augmented Lagrangian \\
& SPCP \citep{zhou2010stable} & RPCA & Accelerated PG \\
& ALM \citep{lin2010augmented} & Both & Augmented Lagrangian \\
\cmidrule(r){1-4}
Matrix & MMMF \citep{rennie2005fast} & MC & Gradient descent \\
factorization & PMF \citep{salakhutdinov2008probabilistic} & MC & Gradient descent \\
& LMaFit \citep{wen2012solving} & MC & Alternating \\
& OptSpace \citep{keshavan2010matrix} & MC & Grassmannian \\
& SET \citep{dai2011subspace} & MC & Grassmannian\\
& LRGeomCG \citep{vandereycken2013low} & MC & Riemannian \\
& GROUSE \citep{balzano2010online} & MC & Online algorithm \\
& JELLYFISH \citep{recht2013parallel} & MC & Stochastic $\&$ parallel \\
& ADMiRA \citep{lee2010admira} & MC & Matching pursuit \\
& SVP \citep{jain2010guaranteed} & MC & Hard thresholding \\
& GRASTA \citep{he2012incremental} & RPCA & Online algorithm \\
& GoDec \citep{zhou2011godec} & RPCA & Hard thresholding \\
& PRMF \citep{wang2012probabilistic} & RPCA & EM algorithm \\
& VB \citep{babacan2012sparse} & Both & Variational Bayes \\
\bottomrule
\end{tabular}}
\begin{tabnote}
\Note{}{This list is inexhaustive. It only includes several representative algorithms in each category.}
\end{tabnote}
\end{table}

\section{Numerical Comparison of Algorithms}\label{sec:comparison}

\hl{

A huge number of solvers have been developed for low-rank matrix recovery in the past decade. \refTab{tab:alg} gives an inexhaustive list of them. In this section, we would like to numerically illustrate their characteristics. As this paper does not focus on a comprehensive comparison, we only test a few solvers on synthesized datasets. Note that the performance of an algorithm often depends on many factors, such as problem size, rank of the underlying low-rank matrix, distribution of its singular values, density of missing entries or outliers, noise level, and even the shape of the matrix. The results presented here only aim to provide a brief demonstration under a few typical settings, which are far from complete. For more detailed comparisons, we refer the readers to the experiment sections in the algorithm papers such as \citep{keshavan2009low,okatani2011efficient,mishra2013fixed} and the report \citep{michenkova2011numerical}. The codes to produce the results presented in this paper is publicly available at \url{https://sites.google.com/site/lowrankmodeling/}. We welcome readers to modify the codes and test the algorithms on more applications.

We evaluate the accuracy of an algorithm by the relative distance defined as
\begin{align}
\mbox{relative distance} = \frac{\|\hat\bfX-\bfX^*\|_F}{\|\bfX^*\|_F},
\end{align}
where $\hat\bfX$ and $\bfX^*$ represent the algorithm estimate and the true low-rank matrix, respectively. As the results depend on stopping conditions of algorithms and different algorithms adopt different stopping criteria, we did not compare the algorithms merely in terms of the relative error of final estimates. Instead, we plot the curve of the relative error for each algorithm as a function of time to see how quickly and how closely the algorithm estimate can approach the ground truth as the algorithm runs. Note that the curves do not aim to show the convergence rates of algorithms since the relative distance here is calculated against the ground truth instead of the stationary point of the cost function. Each curve is averaged over 5 randomly-generated instances with the same problem setting.

The datasets are synthesized in the following way: The low-rank component is a $m\times m$ matrix generated by $\bfX^*=\bfA\bfB^T$, where both $\bfA$ and $\bfB$ are $m\times r$ random matrices with $r\ll m$. $A_{ij}$ and $B_{ij}$ are independently sampled from the normal distribution $\mathcal{N}(0,1)$. Then, $\bfX^*$ is normalized to make $\|\bfX^*\|_F^2=m^2$. For matrix completion, the locations of missing values are defined by a binary matrix with entries independently sampled from the Binomial distribution $\mathcal{B}(1,\rho)$. For RPCA, the locations of outlier entries are simulated in the same way, with their values independently sampled from the uniform distribution $\mathcal{U}(-10,10)$.

For matrix completion, we have tested two convex solvers: the ALM algorithm \citep{lin2010augmented} solving model \refEq{eq:mc_nm} for noiseless cases, and the APG algorithm \citep{toh2010accelerated} solving model \refEq{eq:mcn} for noisy cases. Also, we have tested the following matrix factorization methods: LMaFit \citep{wen2012solving}, OptSpace \citep{keshavan2010matrix}, GROUSE \citep{balzano2010online}, LRGeomCG \citep{vandereycken2013low} and VB \citep{babacan2012sparse}. For RPCA, we have tested two convex models: PCP \citep{candes2011robust} in \refEq{eq:pcp} for noiseless cases, and SPCP \citep{zhou2010stable} in \refEq{eq:spcp} for noisy cases. In addition, we have tested GoDec \citep{zhou2011godec}, PRMF \citep{wang2012probabilistic} and VB \citep{babacan2012sparse} for comparison.

The convex programs were implemented by the authors in MATLAB. For ALM, we implemented the inexact version and adopted the varying penalty parameter scheme provided by \citet{boyd2010distributed}. For APG, we integrated the adaptive restart technique introduced in \citep{o2012adaptive}, which can practically improve the convergence of APG. In original works, partial SVD (only first several singular values are computed) is often used to accelerate the computation for large-scale problems \citep{toh2010accelerated,lin2010augmented}. In our implementation, we simply used the built-in SVD function of MATLAB, since we observed that its efficiency was comparable to or even better than a partial SVD solver (e.g. PROPACK) when $m\leq1000$. SPCP was solved by APG in the original paper \citep{zhou2010stable}. Here, we implemented the block coordinate descent (BCD) algorithm to solve SPCP, i.e., we alternately updated $\bfX$ by SVT and updated $\bfE$ by soft thresholding until convergence. In our experience, the efficiency of BCD is at least comparable to APG to solve the model in \refEq{eq:spcp}, while it is much simpler in implementation. For other algorithms, we used the MATLAB packages downloaded from the authors' websites. We followed the default parameter settings in the original papers and tuned some of them to get reasonably good results for specific problems. Some algorithms require the rank, the noise level or the number of outlier entries as input. To simplify the comparison, we provided their true values to help parameter tuning. We set the initial guess to be the rank-$r$ approximation of the input matrix using SVD for all algorithms.

\subsection{Matrix Completion}

In matrix completion literature, the over-sampling ratio (OS) is widely used to quantify the difficulty of a problem. For a $m\times n$ matrix with rank $r$, the OS is defined as
\begin{align}
\mbox{OS} = |\Omega|/(m+n-r)r,
\end{align}
where $|\Omega|$ denotes the number of observed entries and $(m+n-r)r$ is the underlying degree of freedom of the rank-$r$ matrix. $\mbox{OS}\geq 1$ is required to recover the matrix. The smaller the OS is, the more difficult the recovery turns out to be.

\refFig{fig:mc_acc}(a) shows a basic case, where a sufficient number of entries are observed (OS = 6) and no noise exists. The values of all curves decrease to small numbers below $10^{-6}$, which indicates that all of the algorithms recover the underlying matrix in a high precision. The convex algorithm ALM is comparatively slower, since it needs to perform SVD computation in each iteration. Notice that the convex model in \refEq{eq:mc_nm} is parameter-free, while it recovers the low-rank matrix accurately. The matrix factorization methods are generally accurate and fast. The manifold-based algorithm LRGeomCG achieves the fastest performance, followed by LMaFit. The online algorithm GROUSE also converges to an accurate solution after passing over the data multiple times. The probabilistic method (VB) shows competitive performance while it requires no parameter tuning.

The problem setting in \refFig{fig:mc_acc}(b) is similar to the setting in \refFig{fig:mc_acc}(a) except that the rank is increased from 20 to 50. All curves in \refFig{fig:mc_acc}(b) have similar shapes compared to those in \refFig{fig:mc_acc}(a). The recovery accuracy for all algorithms remains high, as OS is unchanged. However, the curves of the factorization-based methods more or less shift to the right, indicating an increase of computational time. This is attributed to the fact that the dimensions of variables in factorization-based methods directly depend on the predefined rank.

\refFig{fig:mc_acc}(c) shows a case where Gaussian noise with $\sigma=0.1$ is added. The curves of all methods converge to values larger than zero due to the existence of random noise. All of the factorization-based methods achieve similar accuracy, while the relative error of the convex algorithm APG is higher. This is attributed to the fact that, while convex relaxation can guarantee optimality in optimization, it may introduce bias into the model. Specifically, SVT is used to solve the convex model \refEq{eq:mcn}, which will shrink the singular values of the recovered matrix while removing noise components. Consequently, the values of the recovered matrix shrink towards zero, especially when the noise level is large. To compensate for the bias, some postprocessing techniques could be used, which have proven to be effective \cite{mazumder2010spectral}.

\refFig{fig:mc_acc}(d) shows a case where the sampling rate is decreased to OS = 3. Compared to  \refFig{fig:mc_acc}(b) with OS = 6, the curves shift to the right indicating an increase of computational time, which means that the convergence rates of the algorithms are influenced by the over-sampling ratio. Besides, all of the methods still obtain accurate recovery with the relatively low over-sampling ratio.

\begin{figure}
\centering
\begin{minipage}{.47\linewidth}
  \includegraphics[width=\linewidth]{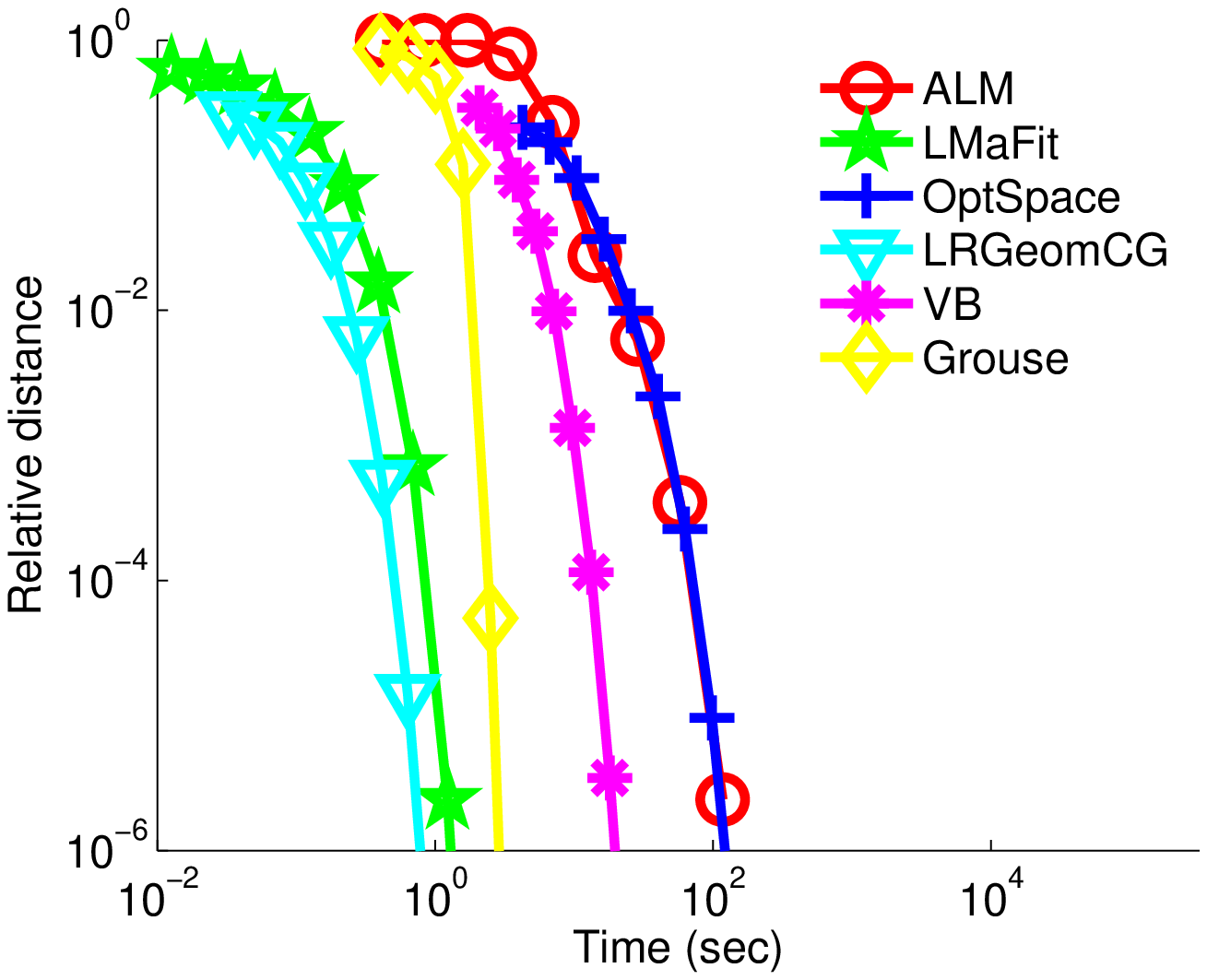}
  \centerline{\small(a) $\mbox{OS}=6$, $r=20$, $\sigma=0$}
\end{minipage}
\begin{minipage}{.47\linewidth}
  \includegraphics[width=\linewidth]{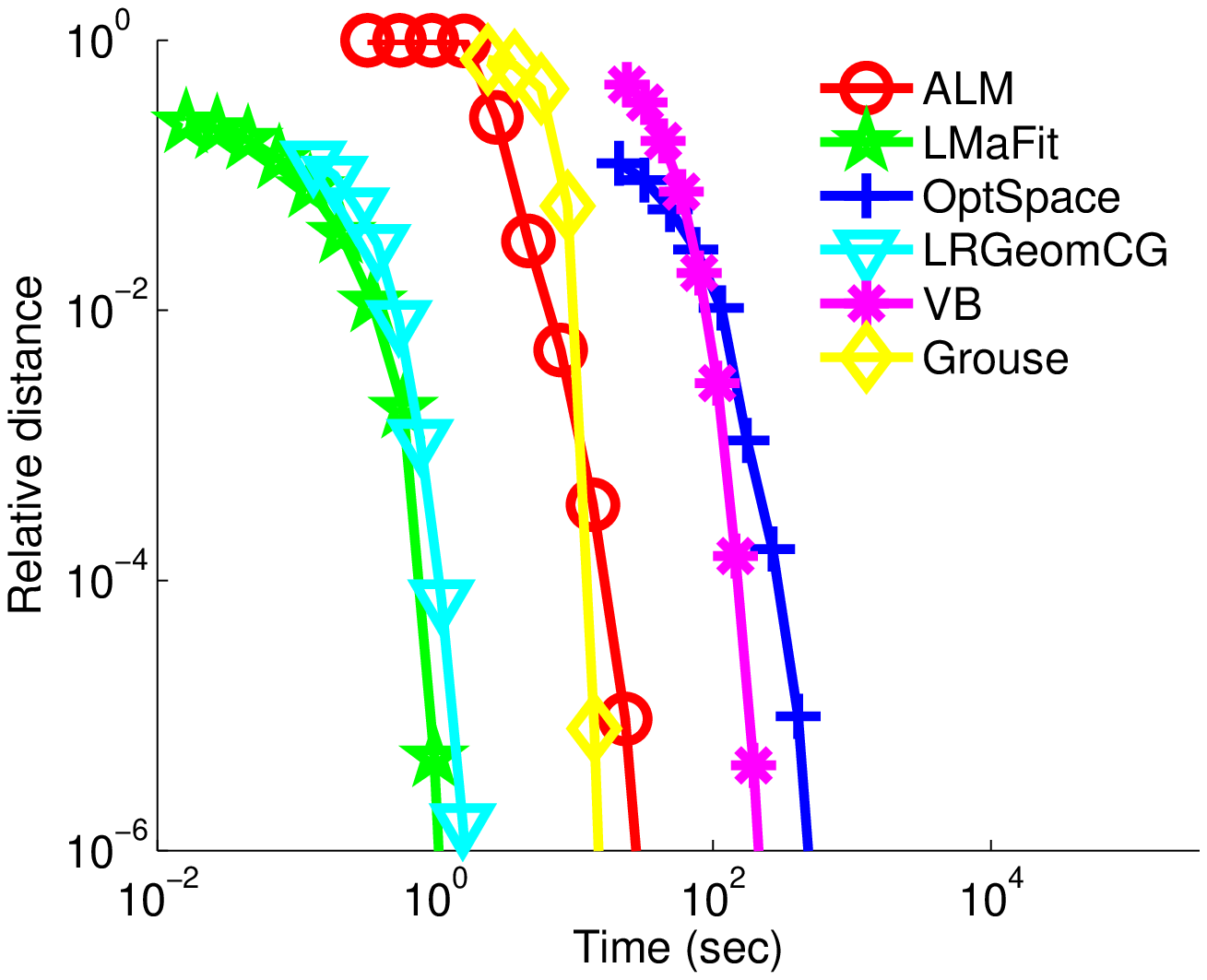}
  \centerline{\small(b) $\mbox{OS}=6$, $r=50$, $\sigma=0$}
\end{minipage}
\begin{minipage}{.47\linewidth}
  \includegraphics[width=\linewidth]{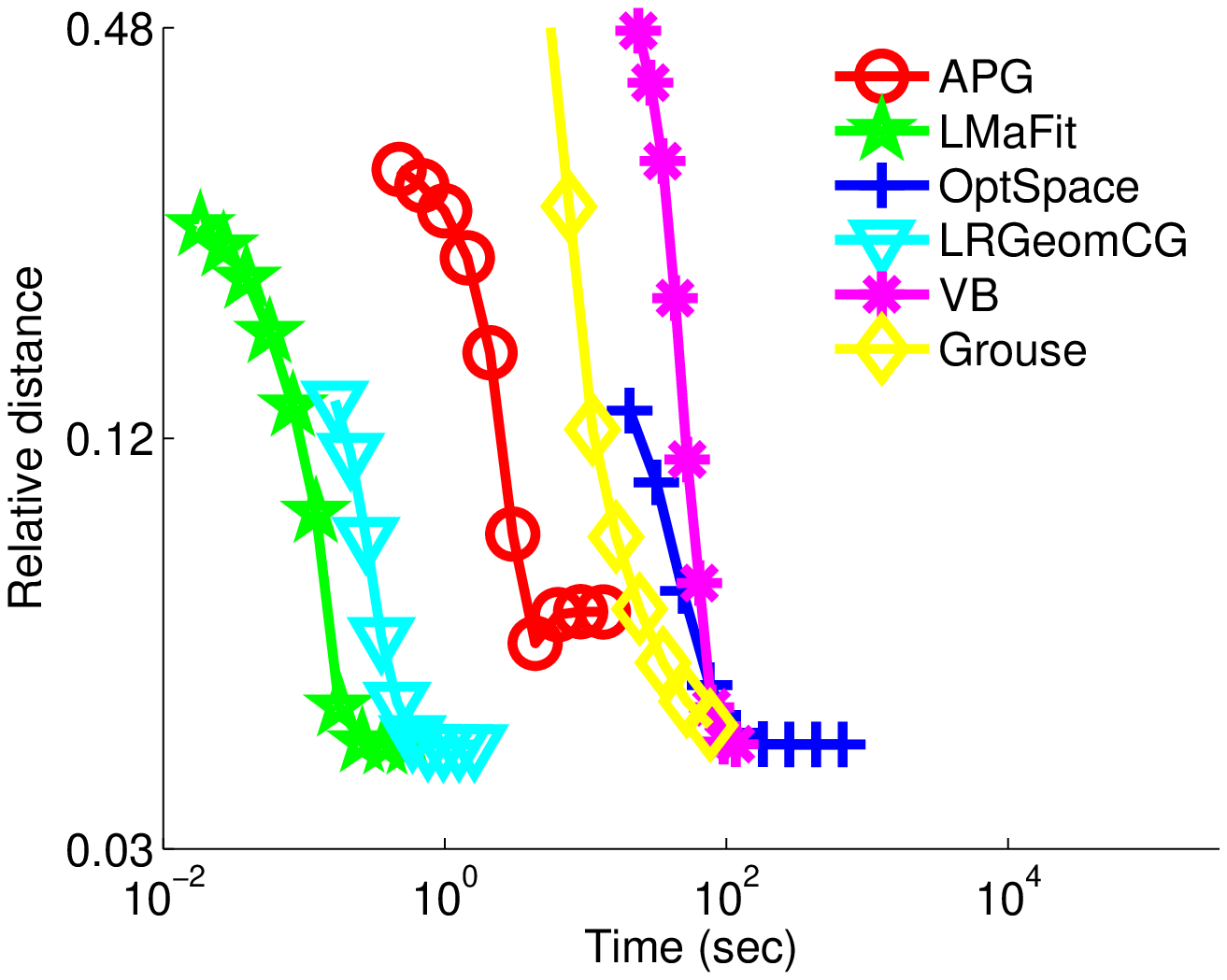}
  \centerline{\small(c) $\mbox{OS}=6$, $r=50$, $\sigma=0.1$}
\end{minipage}
\begin{minipage}{.47\linewidth}
  \includegraphics[width=\linewidth]{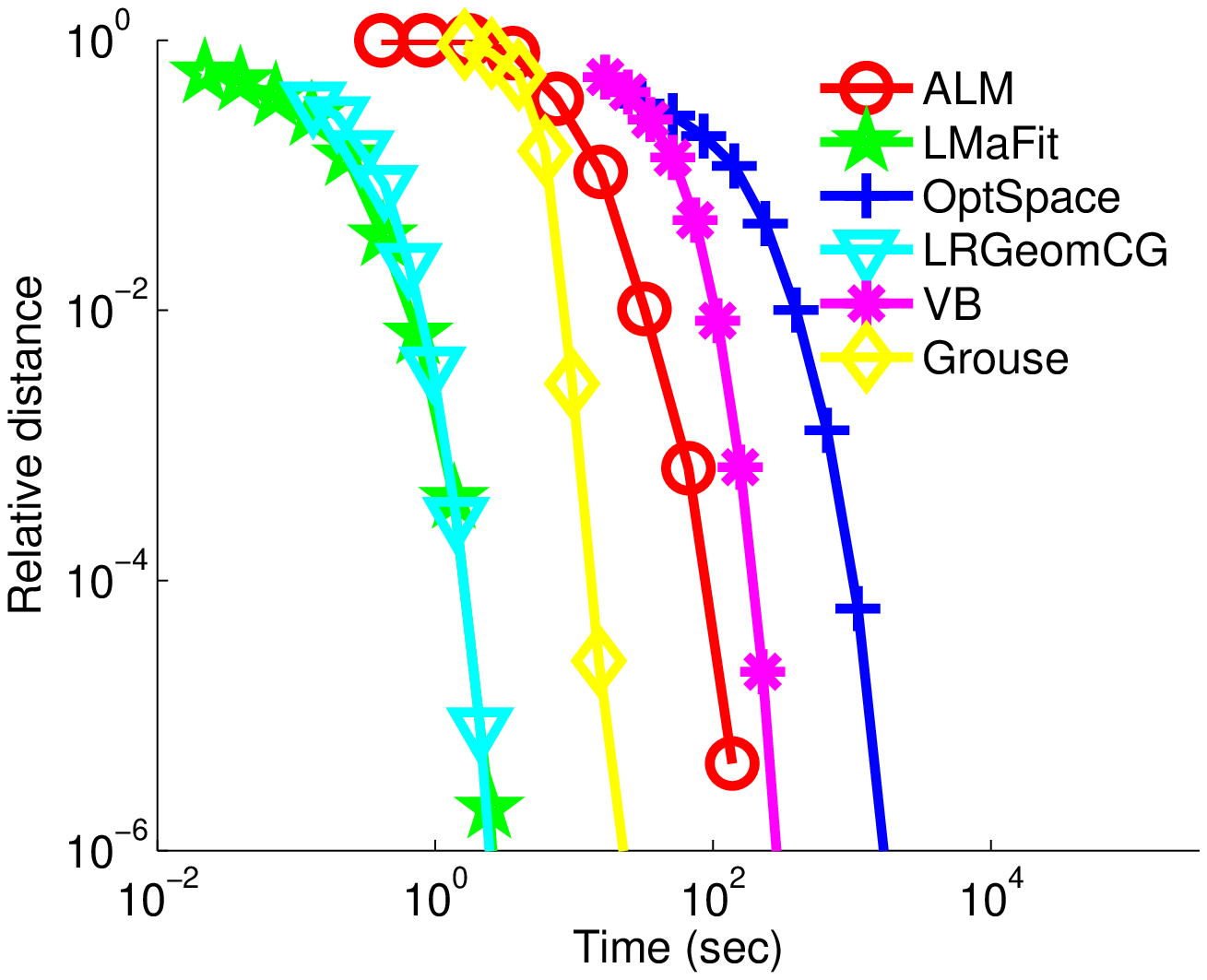}
  \centerline{\small(d) $\mbox{OS}=3$, $r=50$, $\sigma=0$}
\end{minipage}
\caption{Comparison of algorithms for matrix completion. Each curve represents the relative distance between an algorithm estimate and the true low-rank matrix as a function of time on a $\log_{10}/\log_{10}$ scale. The problem size is fixed as $1000\times1000$. OS, $r$ and $\sigma$ denote the over-sampling ratio, the true rank and the noise level, respectively.} \label{fig:mc_acc}
\end{figure}

\subsection{RPCA}

\begin{figure}
\centering
\begin{minipage}{.47\linewidth}
  \includegraphics[width=\linewidth]{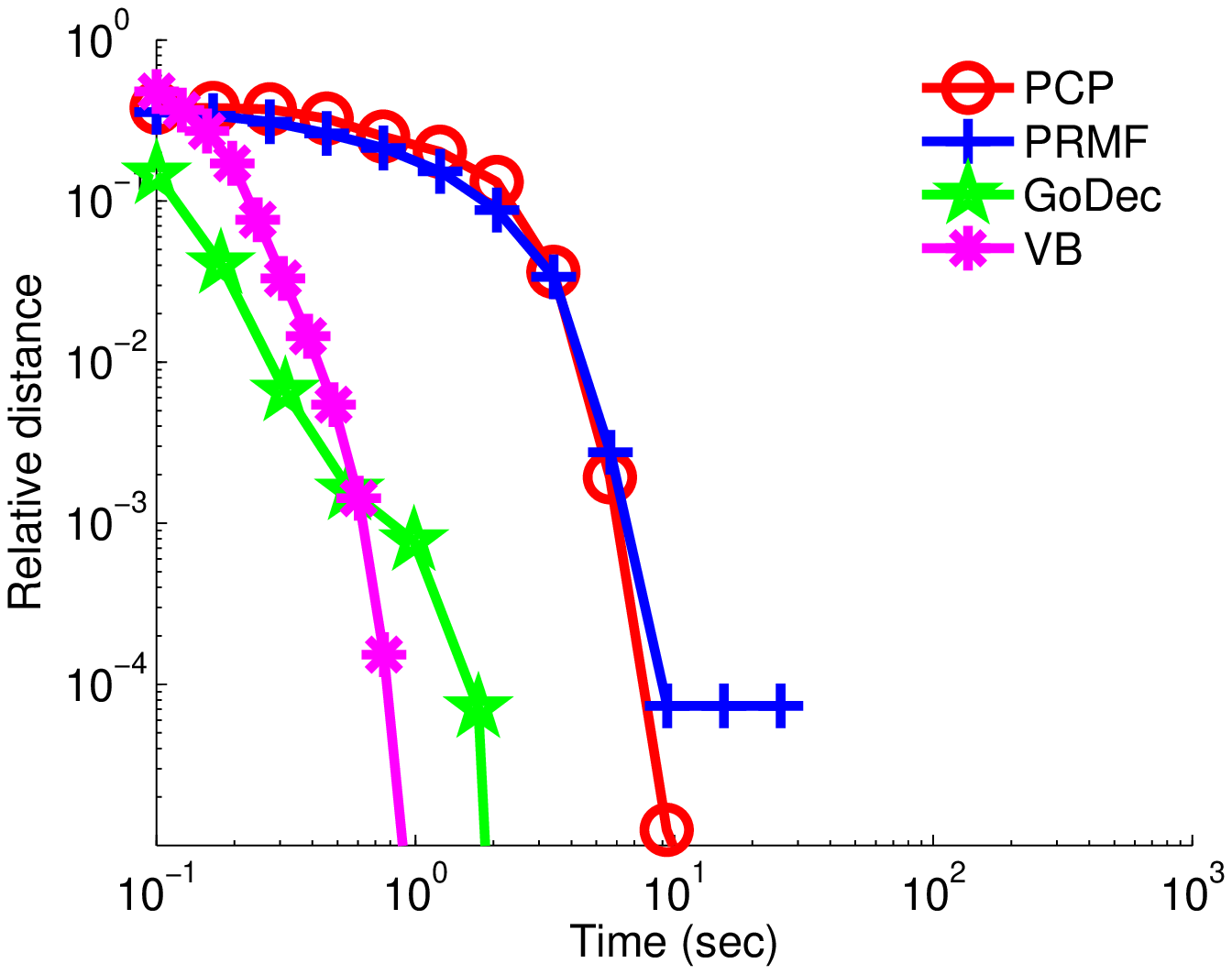}
  \centerline{\small(a). $\rho=0.1$, $r=20$, $\sigma=0$}
\end{minipage}
\begin{minipage}{.47\linewidth}
  \includegraphics[width=\linewidth]{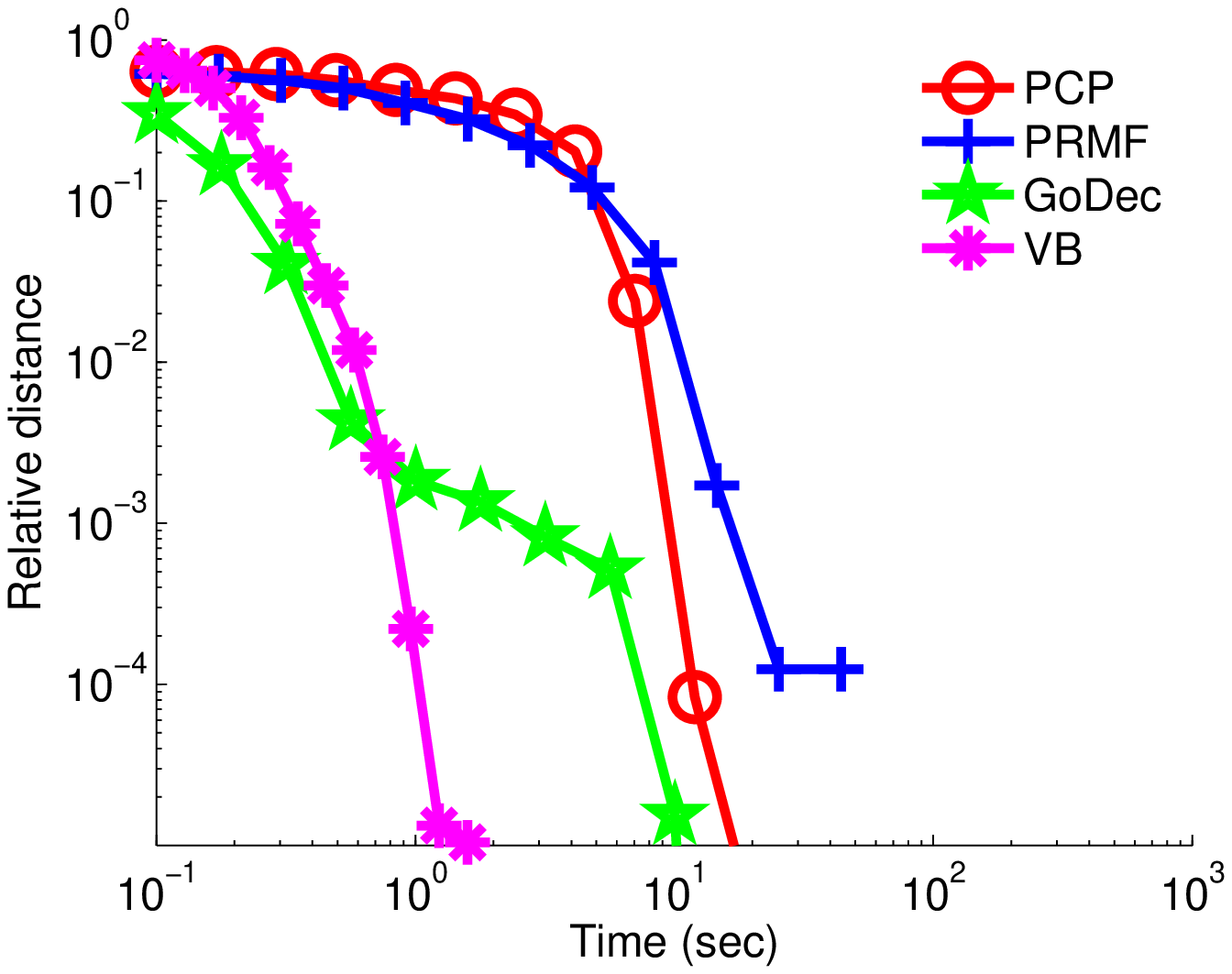}
  \centerline{\small(b). $\rho=0.1$, $r=50$, $\sigma=0$}
\end{minipage}
\begin{minipage}{.47\linewidth}
  \includegraphics[width=\linewidth]{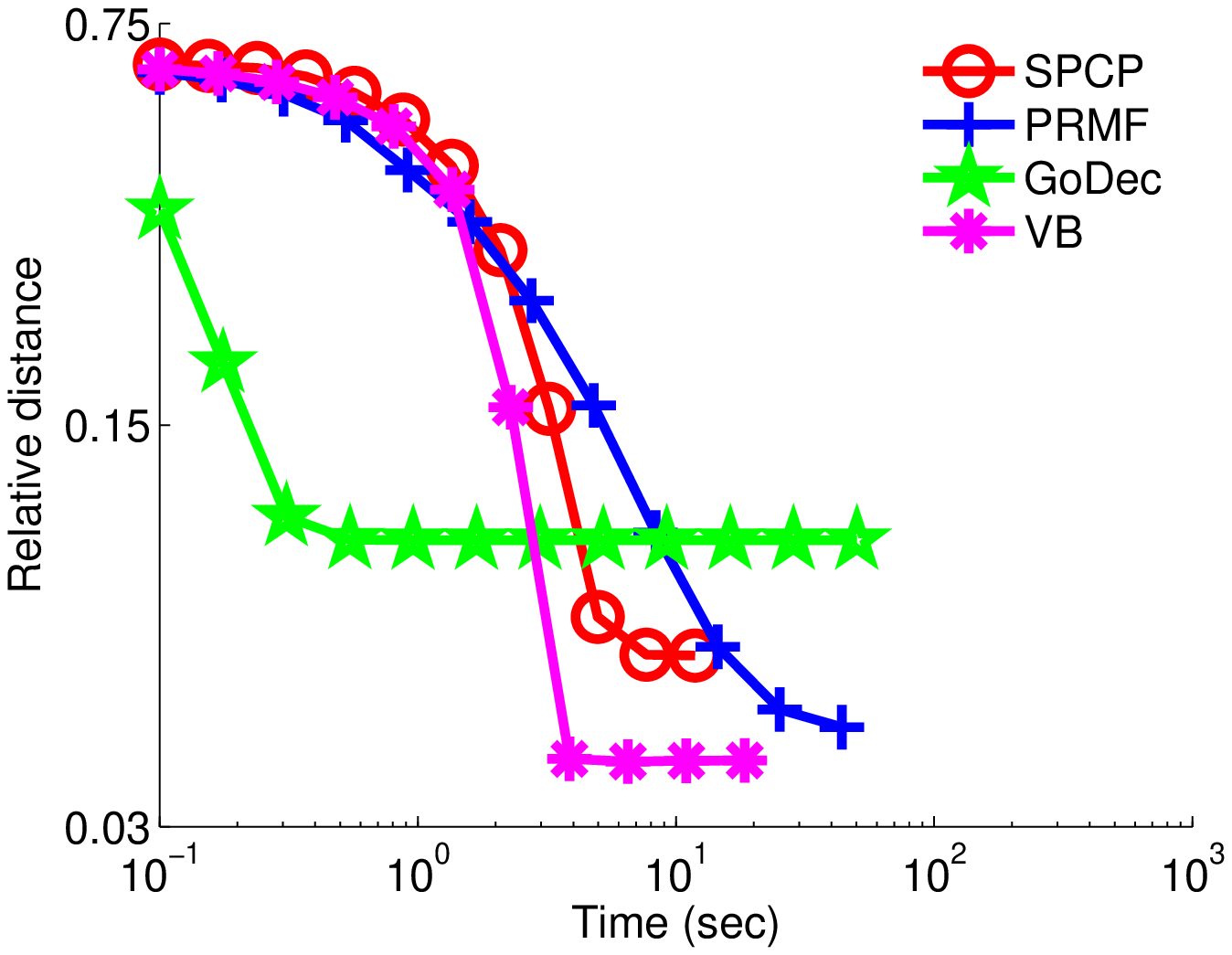}
  \centerline{\small(c). $\rho=0.1$, $r=50$, $\sigma=0.1$}
\end{minipage}
\begin{minipage}{.47\linewidth}
  \includegraphics[width=\linewidth]{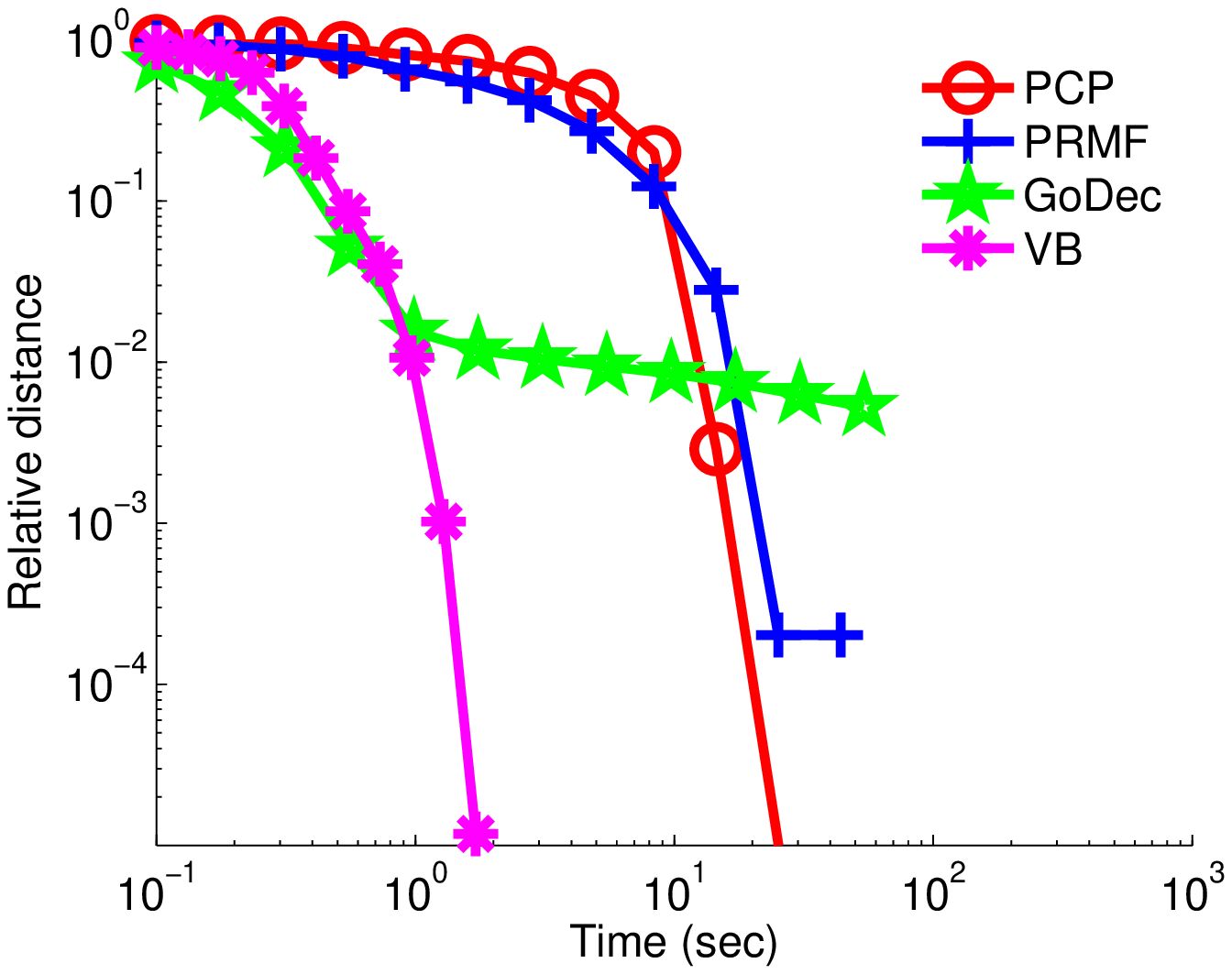}
  \centerline{\small(d). $\rho=0.3$, $r=50$, $\sigma=0$}
\end{minipage}
\caption{Comparison of algorithms for RPCA. Each curve represents the relative distance between an algorithm estimate and the true low-rank matrix as a function of time on a $\log_{10}/\log_{10}$ scale. The problem size is fixed to be $1000\times1000$. $\rho$, $r$ and $\sigma$ denote the proportion of outlier entries, the true rank and the noise level, respectively. } \label{fig:rpca_acc}
\end{figure}

The problem settings for RPCA are similar to the settings for matrix completion. The proportion of outliers is defined as $\rho=|\Omega|/mn$. A larger $\rho$ indicates more outliers, which makes the recovery more difficult. The results are shown in \refFig{fig:rpca_acc}.

The convex program PCP achieves a very high accuracy in noiseless cases without knowing the true rank. For the noisy case, its stable version SPCP is tested, which does not obtain the best accuracy due to the shrinkage effect of nuclear norm minimization as mentioned in the previous subsection. The probabilistic method PRMF achieves similar performance to PCP. Another probabilistic method VB using the Variational Bayes inference achieves the best overall performance in terms of both speed and accuracy, and it requires no parameter tuning. The projection-based method GoDec performs very well under the first two cases, but the performance drops in the more difficult cases in \refFig{fig:rpca_acc}(c) and (d), where the curves are flattened at high values.

}

\section{Applications in Image Analysis}\label{sec:cvapp}

Many objects of interest in image analysis can be modeled as low-rank matrices, such as the images of a convex lambertian surface under various illuminations \citep{basri2003lambertian}, dynamic textures changing periodically \citep{doretto2003dynamic}, active contours with similar shapes \citep{blake2000active}, and multiple feature tracks on a rigid moving object \citep{vidal2004motion}. Intuitively, the low-dimensional subspace models the common patterns underlying the data. Hence, recovering the low-rank structure  is critical to many applications such as background subtraction, face recognition, and segmentation. Below, we will introduce some typical applications based on the models we have discussed in the previous sections.

\subsection{Face Recognition}

\begin{figure}
  \centering
  \includegraphics[width=0.68\linewidth]{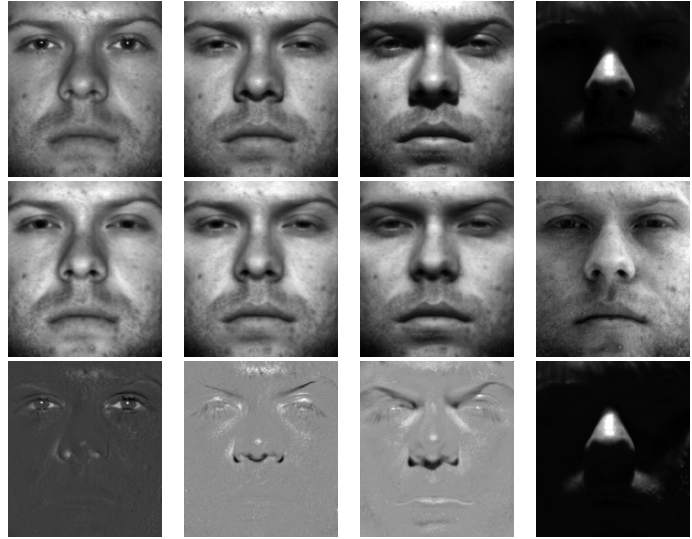}\\
  \caption{Using RPCA to remove shadows and specularities in face images. The rows from top to bottom correspond to the original images, the low-rank components and the sparse components, respectively. We applied the PCP algorithm \citep{candes2011robust} on a set of 64 face images of a person and selected 4 images to show the results. Note that the displayed intensities in each image have been scaled to [0, 255]. Face image courtesy of the Extended Yale Face Database B \citep{georghiades2001few,lee2005acquiring}. }\label{fig:face_rpca}
\end{figure}

The concept of low dimensionality has been used in face recognition for decades since the work by \citet{sirovich1987low}. PCA was applied on a set of face images to construct a face space and each face image can be characterized by a low-dimensional vector \citep{sirovich1987low,kirby1990application}. Later on, \citet{turk1991face} introduced the ``eigenface" method for face recognition. The basic steps of using eigenfaces for face recognition include: (1) generating $N$ eigenfaces by computing the first $N$ eigenvectors of the matrix composed of a set of training images; (2) calculating the weight vector of an input image by projecting the image onto the space spanned by the $N$ eigenfaces; and (3) determining whether the input image is a face image and if so, which person the image belongs to according to the projection error and the weight vector. This is the earliest example of using low-rank modeling for face recognition.

The face images in real datasets are usually corrupted by various artifacts such as shadows, specularities and occlusions, which cannot be handled by classical PCA. Therefore, many approaches based on RPCA were proposed to process face images \citep{de2003framework,candes2011robust,chen2012low}. As illustrated in \refFig{fig:face_rpca}, the local defects in face images could be removed as the sparse component, while the correct description of the person's face could be obtained from the low-rank component. This procedure can improve the characterization of faces and boost the performance of recognition algorithms \citep{chen2012low}.

\subsection{Background Subtraction}

\begin{figure}
  \centering
  \includegraphics[width=0.7\linewidth]{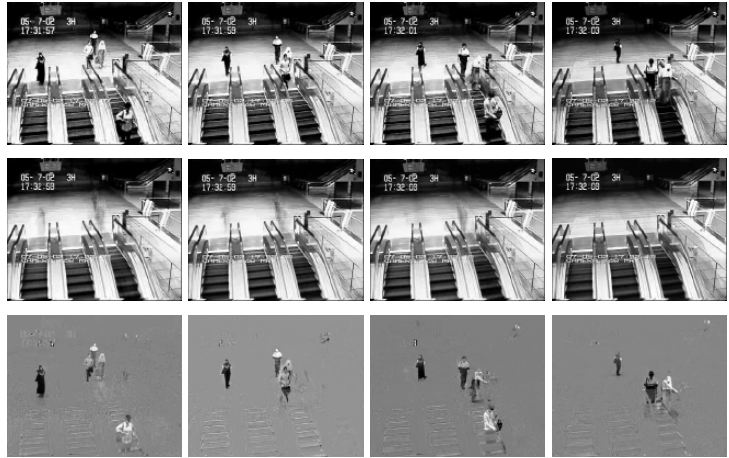}\\
  \caption{Using RPCA for background subtraction. The rows from top to bottom correspond to the input images, the low-rank components (background) and the sparse components (foreground), respectively. We applied the PCP algorithm \citep{candes2011robust} on the subway station dataset from \cite{li2004statistical}. The dataset (used with permission of Liyuan Li) is a surveillance video of a subway station including moving escalators in the background. We selected 200 frames to perform background subtraction and 4 frames to show the results in the figure.}\label{fig:bgs_rpca}
\end{figure}

Background subtraction involves modeling the background in a video and detecting the objects that stand out from the background. Similar to eigenfaces, PCA has been applied to model the background since the work ``eigenbackground subtraction" \citep{oliver2000bayesian}. The basic idea is that the underlying background images of a video captured by a static camera should be unchanged except for illumination variation. Therefore, the matrix composed of vectorized background images can be naturally modeled as a low-rank matrix. However, a set of training images without foreground objects are required to generate a clean background model in traditional methods. To estimate a background model at the presence of foreground objects, RPCA is desired \citep{de2003framework}. As illustrated in \citep{candes2011robust}, the PCP algorithm can recover the background images in the low-rank component and identify the foreground objects in the sparse component. \refFig{fig:bgs_rpca} gives an illustration. Notice that the background includes three moving escalators, which are clearly reconstructed in the low-rank component. This shows the appealing capability of low-rank modeling for background subtraction. To achieve better accuracy for object detection, the spatially-contiguous property of foreground pixels can be modeled and integrated into RPCA by using Markov Random Fields or other smoothing techniques \citep{zhou2012moving,gao2012block,wang2013bayesian}. Similarly, the RPCA framework can be used to segment the point trajectories in a video into two groups, which correspond to background and foreground, respectively \citep{cui2012background}. The segmentation is based on the fact that the background motion caused by camera motion should lie in a low-dimensional subspace.

\subsection{Clustering and Classification}

Low-Rank Representation (LRR) is a well known method for subspace clustering. In subspace clustering, the data points are assumed to be embedded in several low-dimensional subspaces, and the task is to find these subspaces and the membership of each data point in these subspaces. A popular method is spectral clustering, where the clustering is achieved by partitioning a graph whose edge weight represents the affinity between two data points. In LRR, each data point is represented by a linear combination of its neighbors within the same subspace, and the coefficient $\bfX$ is estimated by
\begin{align}\label{eq:lrr}
\min_{\bfX,\bfE} ~& \|\bfX\|_* + \lambda\|\bfE\|_{2,1}, \nonumber \\
\st & \bfD=\bfD\bfX+\bfE,
\end{align}
where $\|\bfE\|_{2,1}=\sum_j\sqrt{\sum_iE_{ij}^2}$ encourages column-wise sparsity on the outlier term $\bfE$. It can be shown that, if the data points in $\bfD$ are from several orthogonal subspaces, $\bfX$ derived from \refEq{eq:lrr} will be block-diagonal \citep{liu2010robust}. Intrinsically, $\bfX$ identifies the affinity between data points, and its block-diagonal structure indicates clusters in the data. Thus, $\bfX$ provides a favorable affinity matrix to perform spectral clustering. A similar idea has also been applied to image segmentation \cite{cheng2011multi}.

Another application of low-rank representation is on dictionary learning for image classification. In dictionary learning, the primary task is to construct a dictionary $\bfPhi=[\phi_i,\cdots,\phi_n]$, such that the input signals can be represented by sparse linear combinations of dictionary atoms. That is, $\bfD=\bfPhi\bfX+\bfE$, where $\bfX$ should be sparse. In \citep{zhang2012low,zhang2013low,zhang2013learning}, it has been claimed that $\bfX$ should also be low-rank to learn a discriminative dictionary. The intuition is that, if the constructed dictionary $\bfPhi$ is discriminative, the signals in $\bfD$ with the same label should be represented by the same set of atoms in $\bfPhi$. Consequently, the coefficient matrix $\bfX$ should be a block-diagonal matrix if the columns are ordered by class labels. To impose such a structural constraint, the sparsity and the rank of $\bfX$ are minimized simultaneously.

\subsection{Image Alignment and Rectification}

Image alignment refers to the problem of transforming different images into the same coordinate system. \citet{peng2011rasl} proposed to solve the problem by rank minimization based on the assumption that a batch of aligned images should form a low-rank matrix. The parameters of transformation $\tau$ were estimated by solving
\begin{align}\label{eq:tilt}
\min_{\tau,\bfX,\bfE}~ &\|\bfX\|_*+\lambda\|\bfE\|_1, \nonumber \\
\st & \bfX+\bfE=\bfD\circ\tau,
\end{align}
where each column of $\bfD$ corresponds to an image to be aligned and $\bfD\circ\tau$ denotes the images after transformation. The sparse component $\bfE$ models local differences among images.

Similarly, \citet{zhang2012tilt} used the model in \refEq{eq:tilt} to generate transform-invariant low-rank textures (TILT). The difference compared to \citep{peng2011rasl} is that $\bfD$ in TILT represents a single image instead of an image sequence. The assumption of TILT is that the rectified images of textures such as characters, bar codes and urban scenes are usually symmetric patterns and consequently form low-rank matrices. The reconstructed low-rank texture can be further used in many applications such as camera calibration, 3D reconstruction, character recognition, etc.

\subsection{Structure and Motion}

The low-rank matrix factorization has been widely used to analyze the tracks of feature points in a video since the seminal work by \citet{tomasi1992shape}. The key observation is that a measurement matrix composed of feature tracks will be rank-limited and the rank depends on the type of camera model (\eg affine or perspective) and the complexity of object motion (\eg rigid or non-rigid). For example, under the weak-perspective model, the 2D image coordinates $\bfx\in\R{2}$ and the 3D position $X\in\R{3}$ of a feature point are related by the following equation:
\begin{align}
\bfx=\bfM\left[
           \begin{array}{c}
             X \\
             1 \\
           \end{array}
         \right],
\end{align}
where $\bfM\in\RR{2}{4}$ is an affine motion matrix. If a set of feature points on a rigid object are tracked across many frames, we have
\begin{align}\label{eq:sfm}
\underbrace{\left[
  \begin{array}{ccc}
    \bfx_{11} & \cdots & \bfx_{1n} \\
    \vdots & \ddots & \vdots \\
    \bfx_{m1} & \cdots & \bfx_{mn} \\
  \end{array}
\right]}_{\bfP\in\RR{2m}{n}} =
\underbrace{\left[
  \begin{array}{c}
    \bfM_1 \\
    \vdots \\
    \bfM_m \\
  \end{array}
\right]}_{\bfM\in\RR{2m}{4}}
\underbrace{\left[
  \begin{array}{ccc}
    X_1 & \cdots & X_n \\
    1 & \cdots & 1 \\
  \end{array}
\right]}_{\bfS\in\RR{4}{n}},
\end{align}
where $\bfx_{ij}$ denotes the 2D image coordinates of point $j$ in frame $i$, $\bfM_i$ is the affine motion matrix for frame $i$, and $X_j$ is the 3D coordinates of point $j$. $\bfP$, $\bfM$ and $\bfS$ are often called measurement matrix, motion matrix and structure matrix, respectively \citep{tomasi1992shape}. Since the smallest dimension of $\bfM$ is 4, we have $\rank{\bfP}\leq 4$. In addition, it is possible to recover the structure and motion matrices from the low-rank factorization of the measurement matrix in \refEq{eq:sfm}, which solves the problem of structure from motion (SFM) \cite{tomasi1992shape}. \hl{For nonrigid SFM, the object shape changes from frame to frame, and \refEq{eq:sfm} is only valid for each frame separately with $m=1$, which is an undetermined system. To make the problem well posed, prior knowledge on the shapes is required. The low-rank prior is widely adopted by many state-of-the-art methods for nonrigid SFM, which assumes that the shapes to be reconstructed are linear combinations of a limited number of basis shapes. Exemplar works include \citep{bregler2000recovering,xiao2006closed,angst2011generalized,dai2012simple}.

Given that the measurement matrix itself is low-rank, the low-rank constraint can also be used to help tracking feature points and finding correspondences across video frames \citep{torresani2002space,irani2002multi,garg2013variational}.}

Another related application is motion segmentation \citep{vidal2004unified,rao2010motion,vidal2011subspace}, where the feature tracks are from multiple moving objects instead of a single rigid object. The task is to segment the feature tracks into different groups, and each group of tracks belong to a single moving object. As discussed above, each group of tracks should form a rank-4 subspace. Therefore, the motion segmentation problem can be formulated as subspace clustering, \ie dividing the feature tracks into multiple clusters with each cluster forming a low-dimensional subspace. For a more detailed introduction to subspace clustering, please refer to \citep{vidal2011subspace}.

\subsection{Restoration and Denoising}

\begin{figure}
  \centering
  \begin{minipage}[b]{0.49\textwidth}
  \includegraphics[width=\linewidth]{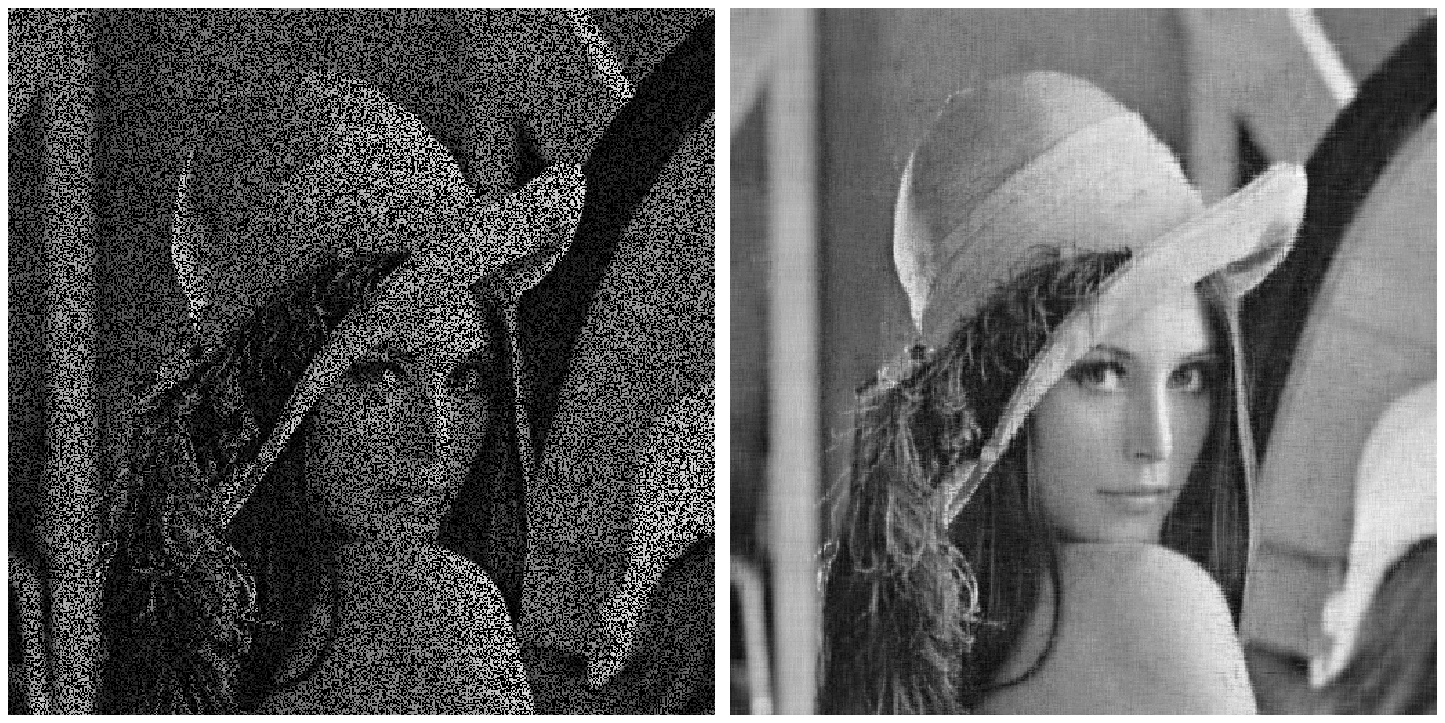} \\
  \centerline{(a)}
  \end{minipage}
  \hfill
  \begin{minipage}[b]{0.49\textwidth}
  \includegraphics[width=\linewidth]{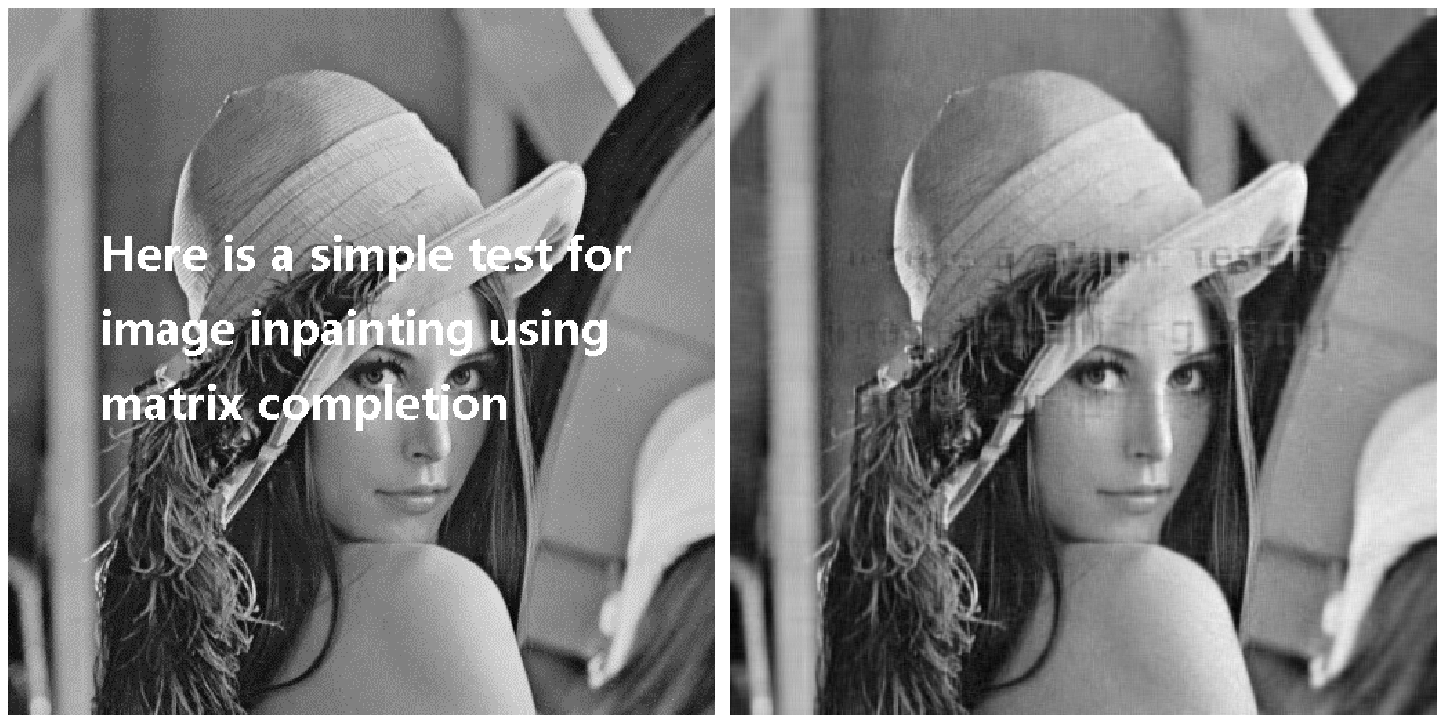} \\
  \centerline{(b)}
  \end{minipage}
  \caption{Matrix completion for image restoration. (a) The input image with $50\%$ missing pixels and the recovered image. (b) The input image corrupted by text and the recovered image. The SOFT-IMPUTE \citep{mazumder2010spectral} algorithm was applied. Original image from \url{http://en.wikipedia.org/wiki/File:Lenna.png}.}\label{fig:inpaint}
\end{figure}

A popular application of matrix completion is image restoration. In many scenarios, it is desired to reconstruct the lost or corrupted parts of an image, which might be caused by texts or logos superposed on the image. This process is named image restoration or inpainting \citep{bertalmio2000image}. As a natural image is approximately low-rank \citep{zhang2012matrix}, the problem of restoring the corrupted pixels can be formulated as a matrix completion problem. \refFig{fig:inpaint} is an illustration of using matrix completion to restore an image from randomly sampled pixels or a text-occluded image. \citet{liang2012repairing} used a more sophisticated model, where the texture to be recovered is modeled as both low-rank and sparse in a certain transformed domain. Moreover, they assumed that the corrupted regions might be unknown and used a sparse error term to model and detect the corrupted regions. To alleviate the shrinkage of signals, some works adopted nonconvex methods for rank minimization instead of using the nuclear norm \citep{zhang2012matrix,gu2014weighted}. \citet{ji2010robust} proposed a method for video restoration. The unreliable pixels in the video are first detected and labeled as missing. Then, the image patches are grouped such that the patches in each group share a similar underlying structure and approximately form a low-rank matrix. Finally, the matrix completion is carried out on each patch group to restore the images.

Similarly, the low-rank assumption is often used to model the coherence of multiple images for noise removal in medical image analysis. In denoising of magnetic resonance (MR) images, for example, an image sequence usually consists of multiple echo images \cite{bydder2006noise}, frames of dynamic imaging \cite{nguyen2011spatiotemporal} or different diffusion-weighted images \cite{lam2012denoising}. Although the images are different, the desired signals in these images are supposed to be correlated and consequently can be reconstructed with several significant principal components. The remaining components correspond to random noise, which are removed. Recently, \citet{candes2012unbiased} used the SVT operator in \refEq{eq:svt} instead of classical PCA to achieve more robust results for image denoising. More importantly, it has been shown that the optimal threshold for SVT can be obtained theoretically based on the Stein's unbiased risk estimate \citep{candes2012unbiased}, which brings great convenience to practical applications.

\subsection{Image Segmentation}

The active shape model \citep{cootes1995active} was proposed to increase the robustness of deformable models for image segmentation. It constructs a statistical shape space from a large set of given shapes and constrains the candidate shape in the shape space. In the active shape model, a candidate shape is represented as
\begin{align}\label{eq:activeshape}
    \mathcal{C}(\bfw) = \overline{\mathcal{C}} + \bfPhi\bfw,
\end{align}
where $\overline{\mathcal{C}}$ denotes the mean shape, $\bfPhi$ is a matrix consisting of vectors describing shape variations in the training data, and $\bfw$ is a vector of coefficients to represent the candidate shape in the shape space. Then, $\bfw$ is determined by fitting the parametric curve in \refEq{eq:activeshape} to the features in the image. Since the number of columns of $\bfPhi$ is often small, the candidate shape is confined in a low-dimensional space. Therefore, the active shape model intrinsically admits a low-rank assumption on the population of shapes. Moreover, $\overline{\mathcal{C}}$ and $\bfPhi$ are derived by applying PCA to the set of training shapes. The active shape model was later extended to the active appearance model to make use of both shape and appearance information \citep{cootes2001active}. More exemplar methods building upon the active shape model for image segmentation include \citep{leventon2000statistical,tsai2001model,cremers2006dynamical,zhu2010segmentation}. An alternative approach to making use of the low-rank assumption for image segmentation is to impose a group-similarity constraint on multiple shapes by nuclear norm minimization \citep{zhou2013active}, which does not require training shapes.

\subsection{Medical Image Reconstruction}
Image reconstruction based on low-rank modeling has drawn much attention in the medical imaging community. The idea is to make use of the temporal coherence in dynamic imaging to reduce the required number of sampling. In MR imaging, for example, \citet{liang2007spatiotemporal} proposed the concept of partial separability (PS) to model a spatial-temporal MR image $\rho(\bfx,t)$ as
\begin{align}\label{eq:psmodel}
    \rho(\bfx,t) = \sum_{\ell=1}^{L}{\phi_{\ell}(\bfx)v_{\ell}(t)},
\end{align}
where $\phi_{\ell}(\bfx)$ and $v_{\ell}(t)$ for $\ell=1,\cdots,L$ are spatial and temporal components, respectively. $L$ is the order of the model. Correspondingly, any sample in the $(\bfk,t)$-space can be expressed as $c(\bfk,t) = \sum_{\ell=1}^{L}{u_{\ell}(\bfk)v_{\ell}(t)}$, where $u_{\ell}(\bfk)$ is the Fourier transform of $\phi_{\ell}(\bfx)$. Using matrix notations, we have
\begin{align}\label{eq:psmatrix}
    \bfC = \bfU\bfV,
\end{align}
where $C_{ij}=c(\bfk_i,t_j)$, $U_{i\ell}=u_{\ell}(\bfk_i)$ and $V_{\ell j}=v_{\ell}(t_j)$. Since the images are temporally coherent, $L$ can be very small, which gives a low-rank model of the coefficients $\bfC$ in the $(\bfk,t)$-space. Hence, a small number of samples are sufficient to estimate $\bfC$ and reconstruct the image sequence. For example, $u_{\ell}(\bfk)$ and $v_{\ell}(t)$ for $\ell=1,\cdots,L$ can be computed by fully sampling $L$ columns and rows of the $(\bfk,t)$-space \citep{liang2007spatiotemporal}. \hl{Some works used other sampling schemes and solved the reconstruction problem by matrix recovery algorithms \citep{haldar2010spatiotemporal,haldar2011low}.}

The basic PS model can be further extended to integrate other sparse properties in specific domains. For example, the image intensity $\phi_{\ell}(\bfx)$ often has a sparse representation in wavelets or a limited total variation \citep{lustig2008compressed,lingala2011accelerated,majumdar2012exploiting}. Meanwhile, the temporal component $v_{\ell}(t)$ is usually periodic or bandlimited, which results in sparsity in the Fourier domain \citep{zhao2010low,zhao2012image}. The low-rank property is modeled as regionally dependent and exploited locally in \citep{christodoulou2012accelerating,trzasko2013exploiting}. Instead of using the PS model, some works \citep{lingala2011accelerated,majumdar2012exploiting} proposed to reconstruct data by nuclear norm minimization. \hl{\citet{otazo2013low} used a RPCA model to separate background and dynamic components in imaging. \citet{lingala2013blind} developed a dictionary learning-based framework for dynamic imaging. Low-rank modeling has also been applied to reconstruct multi-channel data \citep{shin2013calibrationless}, single $\bfk$-space data \citep{haldar2014low} and imaging data from other modalities such as computed tomography (CT) \citep{cai2014cine,gao2011robust} and positron emission tomography (PET) \citep{rahmim2009four}.}

\subsection{More Applications}
Recent examples of low-rank modeling-based applications also include object tracking \citep{ross2008incremental,zhang2012low}, saliency detection \citep{shen2012unified}, correspondence estimation \citep{zeng2012finding,chen2014near}, fa{\c{c}}ade parsing \citep{yang2012parsing}, model fusion \citep{ye2012robust,pan2013divide}, and depth image enhancement \citep{shu2014robust}, to name a few.

\section{Discussions}

In this paper, we have introduced the concept of low-rank modeling and reviewed some representative low-rank models, algorithms and applications in image analysis. \hl{For additional reading on theories, algorithms and applications, the readers are referred to online documents such as the Matrix Factorization Jungle\footnote{URL: \url{https://sites.google.com/site/igorcarron2/matrixfactorizations}. Accessed: 25 May 2014.} and the Sparse and Low-rank Approximation Wiki\footnote{URL: \url{http://ugcs.caltech.edu/~srbecker/wiki/Main_Page}. Accessed: 25 May 2014.}, which are updated on a regular basis.}

The convex programming-based methods for low-rank matrix recovery generally achieve a stable performance under a wide range of scenarios due to the global optimality in optimization. In noiseless cases, the convex programs such as \refEq{eq:mc_rm} and \refEq{eq:pcp} can exactly recover the underlying low-rank matrix with a theoretical guarantee \citep{candes2009exact,candes2011robust}. In noisy cases, the nuclear norm minimization may shrink true signals while compressing noise. To compensate for the shrinkage effect, some postprocessing steps may be used \citep{mazumder2010spectral}, while some other works tried to alleviate this issue by going beyond the nuclear norm and using nonconvex relaxation techniques \citep{mohan2012iterative,zhang2012matrix}. A limitation of convex methods is the requirement of repeated SVD computation, which is time consuming and unaffordable in large-scale problems. While many efforts have been made towards fast SVD computation such as partial SVD \citep{lin2010augmented}, approximate SVD \citep{ma2011fixed} or performing SVT without SVD \citep{cai2010fast}, computational efficiency is still an issue in many real applications.

The factorization-based methods are widely used in real applications (e.g. building recommender systems \citep{koren2009matrix}), mostly due to the computational convenience. Inference of a large matrix is reduced to estimation of two smaller factor matrices. Moreover, the cost function of matrix factorization is often decomposable as a sum of separate functions over data points or variables. Therefore, it is convenient to develop online algorithms for real-time processing and to design distributed algorithms for solving large-scale problems. A limitation of matrix factorization is that the underlying rank needs to be predefined in many models. While rank estimation techniques have been proposed in some works such as \citep{keshavan2009low} and \citep{wen2012solving}, rank estimation is still a challenging problem especially in noisy cases. The probabilistic methods have shown great potential in both simulation \citep{babacan2012sparse} and real application \citep{wang2013bayesian}. Moreover, the probabilistic models with a Bayesian treatment are often parameter free, which is important in real applications.

The applications of low-rank modeling are based on the fact that linear correlation often exists among data. Such prior knowledge can be used for many purposes such as extracting common patterns, removing random noise, reducing sampling rates in imaging, etc. The recent advances in sparse learning and optimization provide powerful frameworks and techniques to conveniently model the low-rank property of data and develop efficient algorithms. We expect to see more applications of low-rank modeling in the near future.

\section*{ACKNOWLEDGMENTS}
The authors thank Michael Klein for carefully proofreading the manuscript.

\bibliographystyle{abbrvnat}
\bibliography{mybib}
\end{document}